\documentclass[10pt,twocolumn,letterpaper]{article}

\usepackage[pagenumbers]{cvpr} 

\usepackage[dvipsnames]{xcolor}

\usepackage[T1]{fontenc}
\usepackage[accsupp]{axessibility} 

\usepackage{arydshln}
\usepackage{booktabs}
\usepackage{enumitem}
\usepackage{balance}
\usepackage{combelow}
\usepackage{tabularx}
\usepackage{cite}
\usepackage{algpseudocode}
\usepackage{amsmath}
\definecolor{azure}{rgb}{0.0, 0.5, 1.0}
\usepackage{wrapfig}
\usepackage[stable]{footmisc}

\usepackage{multirow}
\usepackage{makecell}
\usepackage{mathtools}
\usepackage{algorithm}
\usepackage{graphicx}
\usepackage{algorithmicx}
\usepackage{eqparbox,array}

\usepackage{xcolor,colortbl}
\usepackage{lipsum}
\newcommand\blfootnote[1]{%
  \begingroup
  \renewcommand\thefootnote{}\footnote{#1}%
  \addtocounter{footnote}{-1}%
  \endgroup
}
\usepackage{subcaption}
\usepackage{caption}
\usepackage{bbm}
\usepackage{mwe} 
\usepackage{calc} 
\usepackage{pifont} 

\def\eg{\emph{e.g}\onedot} 
\def\ie{\emph{i.e}\onedot} 
\DeclareMathOperator*{\argmax}{arg\,max}
\DeclareMathOperator*{\argmin}{arg\,min}

\definecolor{azure}{rgb}{0.0, 0.5, 1.0}

\usepackage{subfiles}
\definecolor{cvprblue}{rgb}{0.21,0.49,0.74}
\newcommand{\outref}[1]{\textcolor{blue}{#1}}
\usepackage[pagebackref,breaklinks,colorlinks,citecolor=cvprblue]{hyperref}

\def\paperID{14421} 
\def\confName{CVPR}
\def\confYear{2024}

\title{Learning Equi-angular Representations for Online Continual Learning}

\author{
Minhyuk Seo$^{1, \dagger}$\hspace{0.7em}Hyunseo Koh$^{1}$\hspace{0.7em}Wonje Jeung$^{1}$\hspace{0.7em}Minjae Lee$^{1}$\hspace{0.7em}San Kim$^{1}$\hspace{0.7em}Hankook Lee$^{2,3}$\\
\vspace{0.5em} 
Sungjun Cho$^{2}$\hspace{0.7em}Sungik Choi$^{2}$\hspace{0.7em}Hyunwoo Kim$^{4,*}$\hspace{0.7em}Jonghyun Choi$^{5,*}$\\
$^1$Yonsei Univ.\hspace{0.3em}
$^2$LG AI Research\hspace{0.3em}
$^3$Sungkyunkwan Univ.\hspace{0.3em}
$^4$Zhejiang Lab\hspace{0.3em} 
$^5$Seoul National Univ.\\
{\tt\small {\{dbd0508, specific0924, reccos1020, nasmik419\}@yonsei.ac.kr}} \\
{\tt\small {khs8157@gmail.com, \{sungik.choi,sungjun.cho\}@lgresearch.ai}} \\
{\tt\small {hankook.lee@skku.edu, hwkim@zhejianglab.com, jonghyunchoi@snu.ac.kr}}
}

\begin{document}
\maketitle
\begin{abstract}
    Online continual learning suffers from an underfitted solution due to insufficient training for prompt model update (\eg, single-epoch training).
    To address the challenge, we propose an efficient online continual learning method using the neural collapse phenomenon.
    In particular, we induce neural collapse to form a simplex equiangular tight frame (ETF) structure in the representation space so that the continuously learned model with a single epoch can better fit to the streamed data by proposing preparatory data training and residual correction in the representation space.
    With an extensive set of empirical validations using CIFAR-10/100, TinyImageNet, ImageNet-200, and ImageNet-1K, we show that our proposed method outperforms state-of-the-art methods by a noticeable margin in various online continual learning scenarios such as disjoint and Gaussian scheduled continuous (\ie, boundary-free) data setups.  Code is available at \url{https://github.com/yonseivnl/earl}. \blfootnote{\hspace{-2em}$^\dagger$: Work done while interning at LG AI Research. \\ $~~^*$: Indicates corresponding authors.}
\end{abstract}

\section{Introduction}
\label{sec:introduction}
A growing interest in continuous learning (CL) involves training the model using continuous data streams. 
Mostly, CL research has focused on the \textit{offline} scenario that assumes the model can be trained in multiple epochs for the current task~\citep{rebuffi2017icarl, chaudhry2018riemannian, wu2019large}. 
However, substantial storage and computational complexity are required to store all data to train a model for multiple epochs. 
Recently, there has been significant interest in \emph{online} CL as a more realistic set-up with a lower computational overhead of allowing a single training pass through the data stream~\citep{aljundi2019task, koh2021online, cai2021online}.
Learning a model in streamed data by a single training pass is challenging, since the temporal distribution at each intermediate time point is likely imbalanced, even if the overall distribution of a dataset is balanced.
Imbalanced data distributions would cause several problems, such as bias towards the major classes~\citep{zhao2021energy, kang2021learning} and the hindrance to generalization~\citep{wu2023rethinking}. 

Recently, \emph{minority collapse}~\citep{fang2021exploring}, the phenomenon in which the angles between classifier vectors for minor classes (\ie, the classes that have a relatively small number of samples) become narrow, has been proposed as a fundamental issue in training with imbalanced data, making the classification of minor classes considerably more challenging. 
In contrast, for balanced data sets, it was proven that classifier vectors and the last layer activations for all classes converge into an optimal geometric structure, named the simplex \emph{equiangular tight frame} (ETF) structure, where all pairwise angles between classes are equal and maximally widened when using cross entropy (CE)~\citep{ji2021unconstrained, lu2020neural, zhu2021geometric, wojtowytsch2020emergence} or mean squared error (MSE)~\citep{zhou2022optimization, mixon2020neural, rangamani2022neural, tirer2022extended} loss. 
This phenomenon is called \emph{neural collapse}~\citep{papyan2020prevalence}.

Although neural collapse naturally occurs only in balanced training, several recent studies attempted to induce neural collapse in imbalanced training to address the minority collapse problem using a fixed ETF classifier~\citep{yang2022inducing, zhong2023understanding}. 
When employing a fixed ETF classifier, the output features are pulled toward the corresponding classifier vector during training, since the classifier is fixed.
More recently, research has also been extended to induce neural collapse in offline CL using a fixed ETF classifier~\citep{yang2023neural}.

\begin{figure*}[t!]
    \centering
    \includegraphics[width=\linewidth]{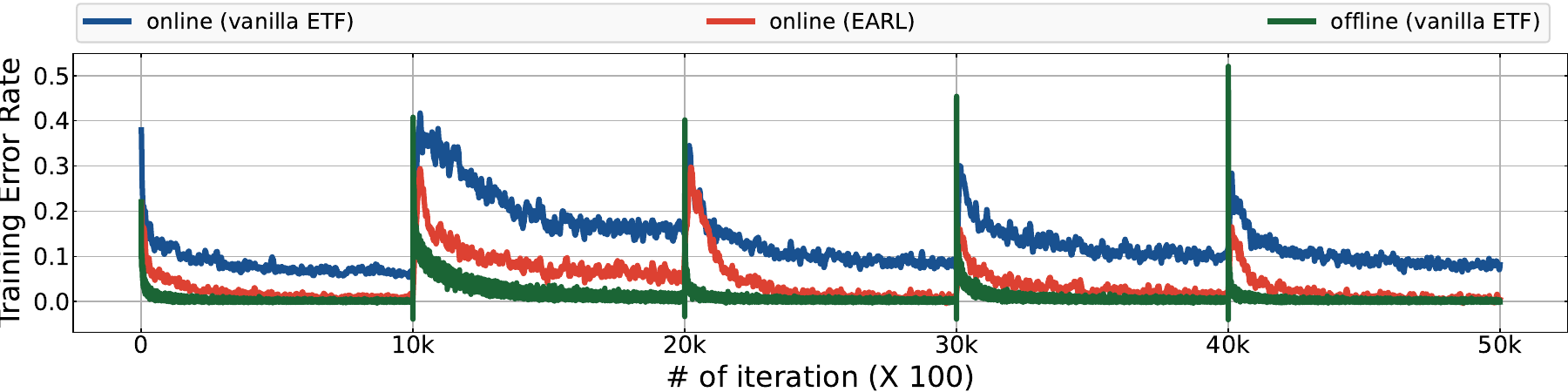}
    \caption{
        Comparison of training error rates between online CL and offline CL in the CIFAR-10 disjoint setup, where two novel classes are added every 10k samples.
        Vanilla ETF refers to a method where both preparatory data training and residual correction are removed from our proposed EARL.
    } 
    \label{fig:tpt}
    \vspace{-1.5em}
\end{figure*}

However, unlike offline CL, there are a number of challenges in inducing neural collapse in online CL, even with a fixed ETF classifier. 
The prerequisite for neural collapse is reaching the \emph{terminal phase of training} (TPT) by sufficient training~\citep{papyan2020prevalence}. 
In offline CL, the model can reach the TPT phase for each task by multi-epoch training. 
In contrast, a single-pass training constraint often prevents the online CL from reaching TPT. 
As shown in Fig.~\ref{fig:tpt}, the offline CL quickly reaches TPT shortly after the arrival of novel task data, while online CL (vanilla ETF) does not.

Recently, the importance of anytime inference in online CL has been emphasized~\citep{pellegrini2020latent, koh2021online, caccia2022anytime, ghunaim2023real}, since a model should be available for inference not only at the end of a task, but also at any point during training to be practical for real-world applications. 
Hence, not only reaching TPT but also achieving faster convergence is necessary when using neural collapse in online CL.

However, we observe that the phenomenon in which the features of the new class become biased toward the features of the existing classes~\cite{caccia2021new} hinders fast convergence of the last-layer features into the ETF structure, which we call the `bias problem.'
When features of old and new classes overlap (\ie, biased) and are trained with the same objective, well-clustered features of old classes disperse (or perturb)~\cite{caccia2021new}, leading to destruction of the ETF structure formed by features of the old classes.

\vspace{-1em}
\paragraph{Contributions.} 
To address the bias problem, we train the model to distinguish out-of-distribution (OOD) samples from existing classes' samples.
Specifically, we synthesize \emph{preparatory data} that play the role of OOD data, by applying negative data augmentation~\cite{sinha2021negative, wang2022resmooth, kim2022continual} to samples from existing classes, refer to Sec.~\ref{sec:preparatory} for details.
Since samples from new classes are OOD in the perspective of existing classes, training with preparatory data encourages the representation of new classes to be distinguished from existing classes in advance, thereby mitigating the bias problem.
This promotes fast and stable convergence into the ETF structure.

Despite these efforts, however, the continuous stream of new samples cause ongoing data distribution shifts, which prevent the model from reaching the TPT and fully converging to the ETF structure. 
To address this, we propose to store the residuals between the target ETF classifier and the features during training.
During inference, we correct the inference output using the stored residual to compensate for insufficient convergence in the training.

By accelerating convergence by preparatory data training and additional correction using residual, we noticeably improve anytime inference performance than state of the arts.
We name our method \textbf{Equi-Angular Representation Learning} (\textbf{EARL}).
We empirically demonstrate the effectiveness of our framework on CIFAR-10, CIFAR-100, TinyImageNet, Imagenet-200, and ImageNet-1K. In particular, our framework outperforms various CL methods by a significant margin (+4.0\% gain of $A_\text{auc}$ in ImageNet-1K).

We summarize our contributions as follows:
\begin{itemize}[leftmargin=15pt]
\setlength\itemsep{0.1em}
    \item Proposing to induce neural collapse for online CL.
    \item Proposing `preparatory data training' to address the `bias problem' that the new classes are biased toward the existing classes, promoting faster induction of neural collapse.
    \item Proposing `residual correction' scheme to compensate for not fully reaching neural collapse at inference to further improve anytime inference accuracy.
\end{itemize}

\section{Related work}

\paragraph{Continual learning methods.}
Various continual learning methods are being researched to prevent forgetting past tasks, broadly categorized into replay, parameter isolation, and regularization.
Replay methods involve storing a small portion of data from previous tasks in episodic memory~\citep{hayes2020remind, aljundi2019task, koh2021online, bang2021rainbow, yoon2021online} or storing a generative model trained on data from previous tasks~\citep{shin2017continual, pomponi2023continual}. 
By replaying the samples stored in episodic memory or generated from the stored generative model, the model prevents forgetting past tasks during subsequent learning of novel tasks. 
Furthermore, ~\citet{boschini2022class, buzzega2020dark, li2017learning, wu2019large} used replay samples to distill information about past tasks.

Regularization methods~\citep{kirkpatrick2017overcoming, lesort2019regularization} apply a penalty to the change of important model parameters during the process of learning new tasks, allowing the model to retain information about previous tasks.
Parameter isolation methods~\citep{zhou2022model, rusu2016progressive, cheung2019superposition} expand the network by allocating specific layers for each task. This enables the network to store information about individual tasks and preserves their knowledge without forgetting.

\vspace{-1em}
\paragraph{Neural collapse.}
Neural collapse is a phenomenon in which the activations of the last layer and the classifier vectors form a simplex \emph{equiangular tight frame} (ETF) structure at the terminal phase of training (TPT) in a balanced dataset.~\citep{papyan2020prevalence}. 
Neural collapse has been demonstrated as the global optimum of balanced training using cross entropy (CE) loss~\citep{ji2021unconstrained, lu2020neural, zhu2021geometric, wojtowytsch2020emergence} and MSE~\citep{zhou2022optimization, mixon2020neural, rangamani2022neural, tirer2022extended} loss functions, within a simplified model focused solely on last-layer optimization. Inducing neural collapse in imbalanced datasets poses challenges due to minority collapse~\citep{fang2021exploring} where minor classes are not well distinguished. 

However, using a fixed ETF classifier, it is empirically and theoretically shown that neural collapse is induced even in imbalanced datasets~\citep{yang2022inducing}.
Continual learning also needs to address imbalanced data since there is an imbalance in the data between novel classes and existing classes. 
Therefore, in offline CL, NC-FSCIL~\citep{pernici2021class, yang2023neural} used a fixed ETF classifier to induce neural collapse.
Meanwhile, online CL often fails to induce neural collapse compared to offline CL, \eg, FSCIL, since it lacks sufficient training in multiple epochs, causing failure to reach TPT.

Please refer to the supplementary material for more comprehensive literature reviews including topics related to anytime inference.

\section{Preliminaries}
We here describe the problem statement for the online continual learning (Sec.~\ref{sec:ps_ocl}), 
neural collapse (Sec.~\ref{sec:nc_define}), and the equiangular tight frame classifier (Sec.~\ref{sec:etf_classifier}). 

\subsection{Problem Statement of Online CL}
\label{sec:ps_ocl}
Continual learning (CL) aims to learn from a stream of data, rather than a fixed dataset as in standard learning.
Formally, given a sequence of tasks $\mathcal{T} = (T_1, T_2, \ldots)$ where each task $T_i$ is a training dataset 
$D_i=\{(x^{(i)}_1, y^{(i)}_1), (x^{(i)}_2, y^{(i)}_2),\ldots\}$,
an offline CL algorithm $A_\text{CL}$ updates the model parameter $\theta$ based on the current task $D_k$, \ie, $\theta_k = A_\text{CL}(\theta_{k-1}, D_k)$, starting from the initial parameter $\theta_0$.

To avoid the forgetting issue in CL, many CL setups allow the use of episodic memory $M_k$, which is a limited-size subset of training data from previous tasks, \ie, $(\theta_k, M_k) = A_\text{CL}(\theta_{k-1}, M_{k-1}, D_k)$.
The objective is to minimize the error of $\theta_k$ in all observed tasks $\{T_i\}_{i=1}^k$. 

Unlike offline CL where all training samples in $D_k$ are given as input, in online CL, the input is provided as a stream of samples $(x^{(k)}_1, y^{(k)}_1), (x^{(k)}_2, y^{(k)}_2),\cdots$.
Therefore, an online CL algorithm $A_\text{OCL}$ is defined as: 
\begin{equation}
(\theta_{k, t}, M_{k, t}) = A_\text{OCL}\left(\theta_{k, t-1}, M_{k, t-1}, (x^{(k)}_t, y^{(k)}_t)\right),
\end{equation}
with the same objective of minimizing the error of $\theta_{k, t}$ on $\{T_i\}_{i=1}^k$. 
Since the algorithm does not have access to previous samples $\{(x^{(k)}_{s}, y^{(k)}_{s}),s<t\}$ at time $t$, multi-epoch training with $D_k$ is unavailable in online CL. 



\subsection{Neural Collapse}
\label{sec:nc_define}
Neural collapse~\citep{papyan2020prevalence} is a phenomenon of the penultimate features after the convergence of training on a balanced dataset. 
When neural collapse (NC) occurs, the collection of $K$ classifier vectors $\mathbf{W}_\text{ETF} = [\mathbf{w}_1, \mathbf{w}_2, ..., \mathbf{w}_K] \in \mathbb{R}^{d \times K}$ forms a simplex equiangular tight frame (ETF), which satisfies:
\begin{equation}
    \mathbf{w}_i^T \mathbf{w}_j = 
    \begin{cases}
        ~1, ~~~~~~~~~~~~~~~~ i=j \\ 
        -\frac{1}{K-1}, ~~~~~~~~i \neq j
    \end{cases},
    ~~\forall i, j \in [1,..., K],
\end{equation}
where $K-1\leq d$, and the penultimate feature of a training sample collapses into an ETF vector $\mathbf{w}_i$.





\subsection{Equiangular Tight Frame (ETF) Classifier}
\label{sec:etf_classifier}

Inspired by neural collapse as described in Sec.~\ref{sec:nc_define}, a fixed ETF classifier has been utilized for inducing neural collapse in imbalanced datasets \citep{zhu2021geometric, yang2022inducing, yang2023neural}.
Here, the classifier is initialized by the ETF structure $\mathbf{W}_\text{ETF}$ at the beginning of training and fixed during training to induce the penultimate feature $f(x)$ to converge to the ideal balanced scenario.
For training $f(x)$, it is only required to attract $f(x)$ to the corresponding classifier vector for convergence, since the classifier is fixed during training. 
Therefore, following \citet{yang2022inducing}, we use the \emph{dot regression} (DR) loss as a training objective, as it shows to outperform cross entropy (CE) loss when using a fixed ETF classifier in imbalanced datasets~\citep{yang2022inducing}.\footnote{If the training objective includes a contrastive term between different classes like cross entropy, it could cause incorrect gradients~\citep{yang2022inducing}.} The DR loss can be written as follows:
\begin{equation}
    \mathcal{L}_\text{DR}(\hat{f}(x),y;\mathbf{W_{\text{ETF}}})=\frac1{2}\Big (\mathbf{w}_{y}\hat{f}(x)-1\Big )^2 ,
\end{equation}
where $\hat{f}(x)= f(x) / \|f(x)\|_2$ is the $L_2$ normalized feature of the model $f$, $y$ is the label of the input $x$, and $\mathbf{w}_{y}$ is a classifier vector in $\mathbf{W}_\text{ETF}$ for the label $y$.

\section{Approach}

\begin{figure*}[t!]
    \includegraphics[width=\linewidth]{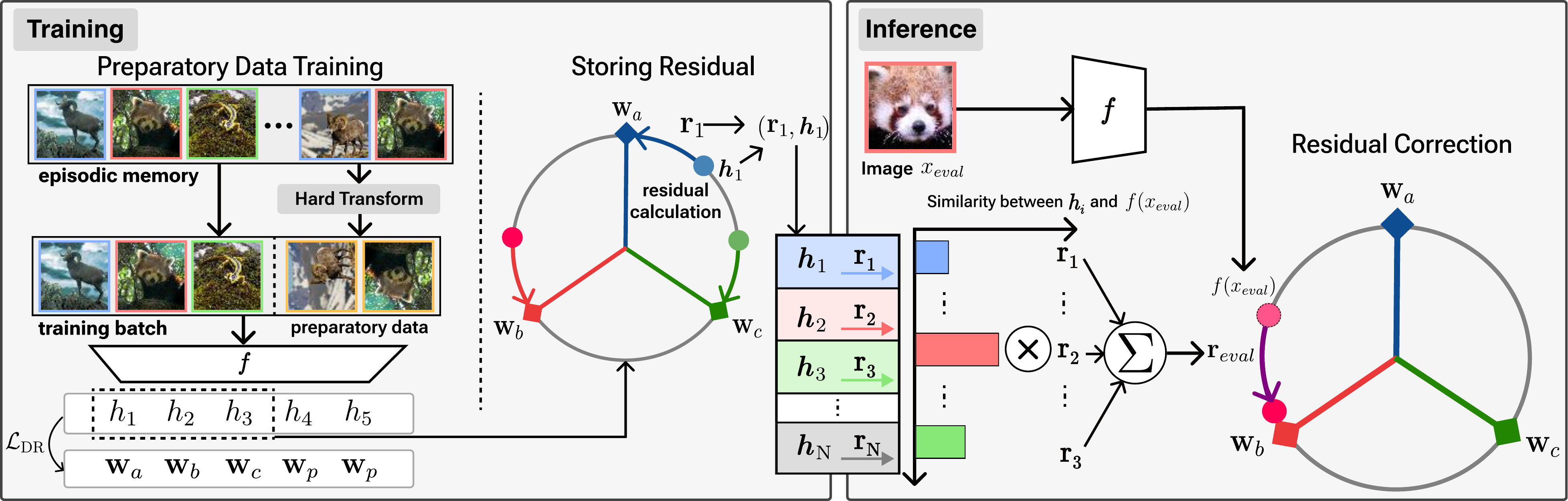}
    \vspace{-1.5em}
    \caption{Overview of EARL. $\mathbf{w}_i$ denotes the ETF classifier vector for class $i$. $h_a$ denotes the output of the model. The colors of the data denote the class to which the data belong. The arrow $\mathbf{r}_i$ denotes the residual between the last layer activation $h_i$ and the classifier vector $\mathbf{w}_i$ for class $i$.
    During training, both memory and preparatory data are used for replaying, and the residuals between $h_i$ and $\mathbf{w}_i$ are stored in feature-residual memory.
    During inference, using the similarity between $f(x_\text{eval})$ and $h_i$ in feature-residual memory, $\textbf{r}_\text{eval}$ is obtained by a weighted sum of $\mathbf{r}_i$'s. Finally, by adding $\mathbf{r}_\text{eval}$, $f(x_\text{eval})$ is corrected. The purple arrow indicates `residual correction' (Sec.~\ref{sec:residual}).
    } 
    \label{fig:overview_setup}
    \vspace{-0.5em}
\end{figure*}

\label{sec:approach}
Despite the success of the fixed ETF classifier in both imbalanced training~\citep{yang2022inducing, zhong2023understanding} and offline CL~\citep{yang2023neural}, the ETF classifier has not yet been explored for online CL due to the necessity of sufficient training for neural collapse.
To be specific, streamed data are trained only once in online CL, which makes it harder to induce neural collapse than in offline CL which supports multi-epoch training.

\begin{figure}[t!]
    \centering
    \begin{subfigure}[b]{1\linewidth}
        \centering
        \resizebox{0.9\linewidth}{!}{
            \includegraphics{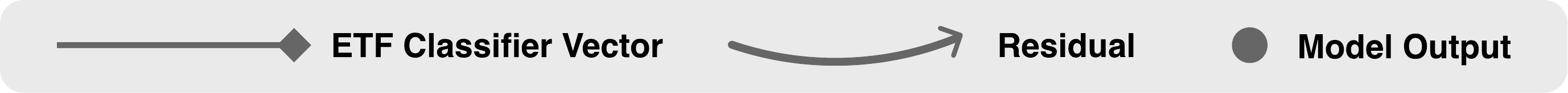}
        }
        \vspace{0.5em}
    \end{subfigure}
    \begin{subfigure}[b]{0.32\linewidth}
        \centering
        \resizebox{0.9\linewidth}{!}{
            \includegraphics{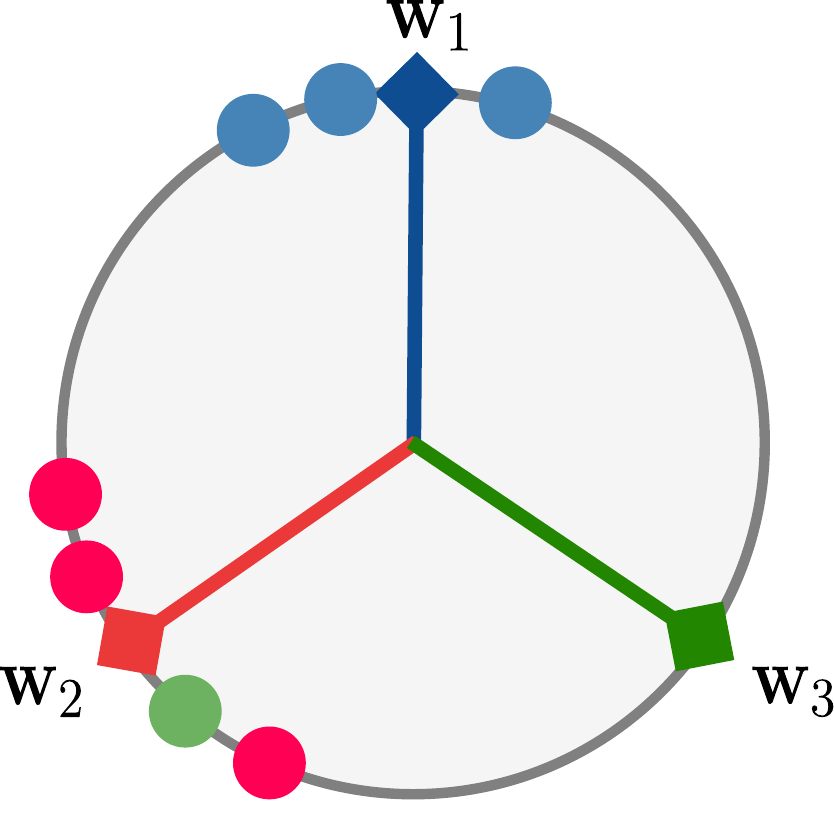}
        }
        \vspace{-0.1em}
        \caption{Vanilla ETF}
    \end{subfigure}
    \begin{subfigure}[b]{0.32\linewidth}
        \centering
        \resizebox{0.9\linewidth}{!}{
            \includegraphics{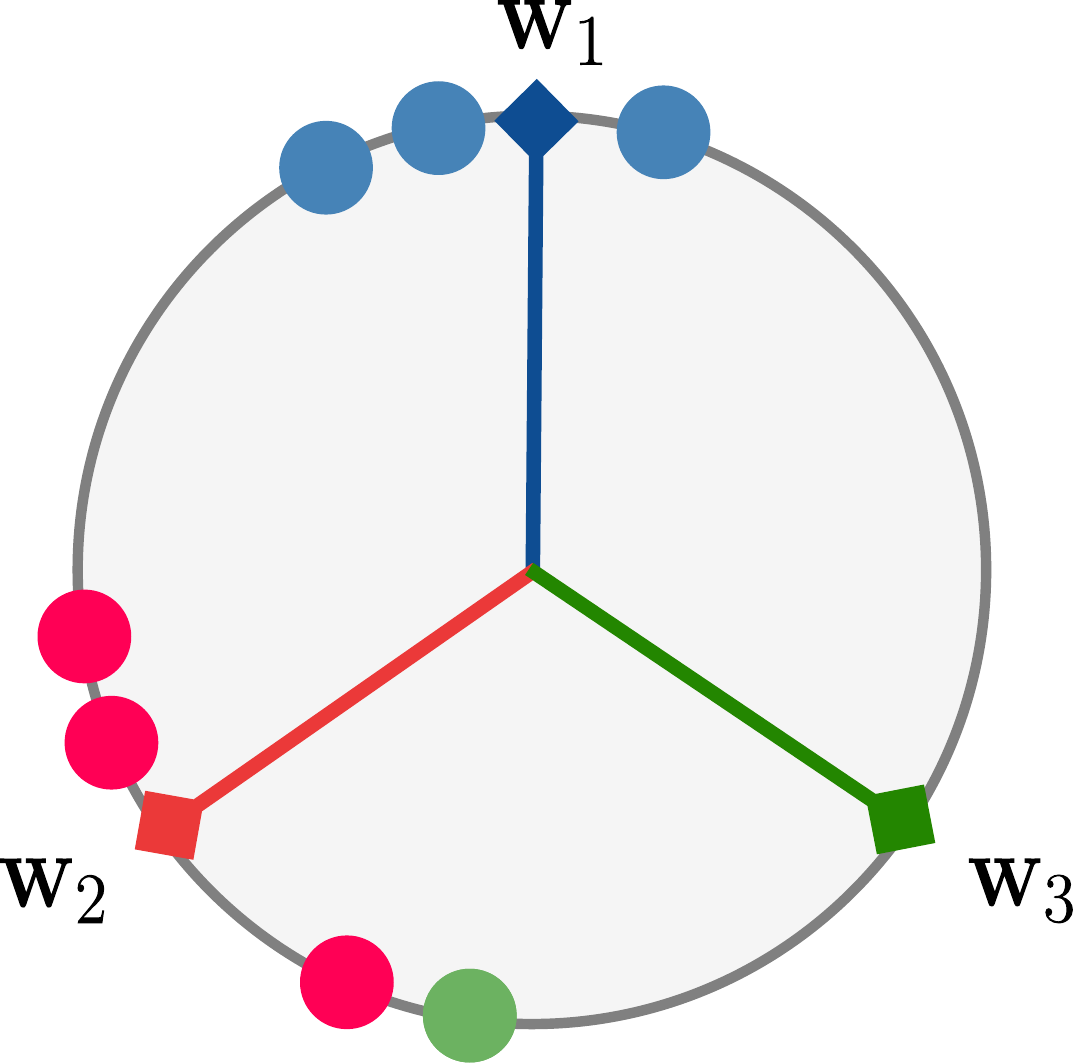}
        }
        \vspace{-0.1em}
        \caption{w/ Prep. Data} 
    \end{subfigure}
    \begin{subfigure}[b]{0.32\linewidth}
        \centering
        \resizebox{0.9\linewidth}{!}{
            \includegraphics{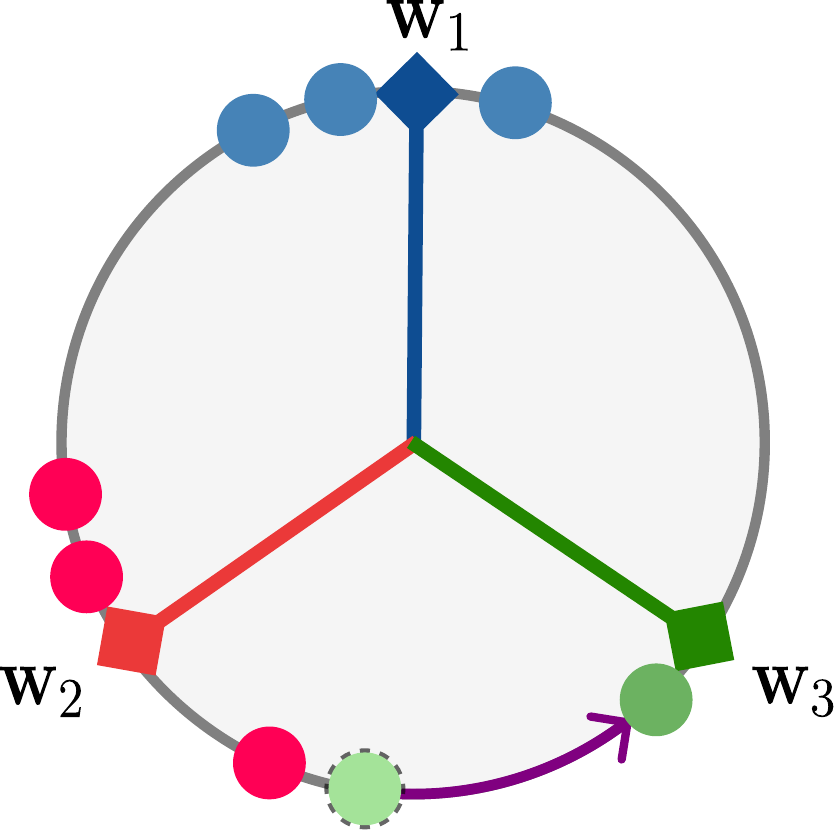}
        }
        \vspace{-0.1em}
        \caption{w/ Res. Corr.}
    \end{subfigure}
    \vspace{-0.5em}
    \caption{Illustrative effects for each component of EARL. (a) In online CL, features of novel classes are biased towards the features of the previous class. 
    (b) By training with preparatory data (\textbf{w/ Prep. Data}, Sec.~\ref{sec:preparatory}), we address the bias problem. 
    (c) In inference, for features that do not fully converge to an ETF classifier, we add residuals (\textbf{w/ Res. Corr.}, Sec.~\ref{sec:residual}) to features that have not yet reached the corresponding classifier vectors, making features aligned with them. Purple arrow: the `residual correction', Colors: classes.}
    \vspace{-1.5em}
    \label{fig:effect}
\end{figure}

To learn a better converged model without multi-epoch training for online CL, we propose two novel methods, each for the training phase and the inference phase, respectively.
In the training phase, we accelerate convergence by proposing \emph{preparatory data training} (Sec.~\ref{sec:preparatory}).
In the inference phase, we propose to correct the remaining discrepancy between the classifiers and the features using \emph{residual correction} (Sec.~\ref{sec:residual}). 
We illustrate an overview of EARL in Fig.~\ref{fig:overview_setup} and the effect of EARL in Fig.~\ref{fig:effect}. 

\subsection{Inducing Neural Collapse for Online CL}
\label{sec:etf_baseline}
For online CL, we first try to induce neural collapse with a fixed ETF classifier: $\mathbf{W}_\text{ETF} \in R^{d \times K}$ where $d$ is the dimension of the embedding space and $K$ is the number of ETF vectors. 
While prior works~\cite{yang2022inducing, yang2023neural} use prior knowledge of the total number of classes that will be encountered for setting $K$, it is unrealistic to know the knowledge (\ie, the exact number of classes) as it evolves continuously over time under realistic CL scenarios. 

To address the challenge, we propose using the maximum possible number of classifier vectors for $K$. Based on the simplex ETF property that the maximum number of ETF vectors is $d+1$~\cite{fickus2018equiangular, yang2022inducing}, we define our ETF classifier $\mathbf{W}_\text{ETF} \in \mathbb{R}^{d \times (d+1)}$ by letting $K:=d+1$. 

\subsection{Preparatory data at training}
\label{sec:preparatory}

\begin{figure}[t!]
    \vspace{0em}
    \centering
    \includegraphics[width=1.0\linewidth]{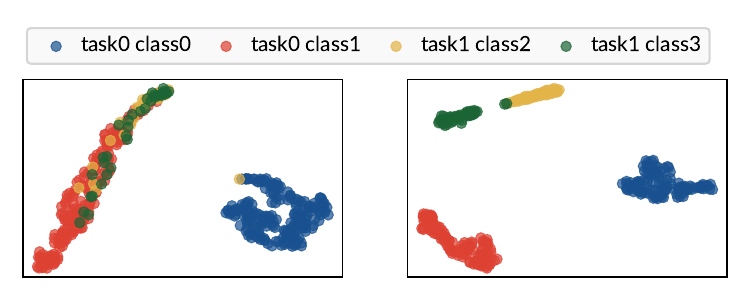}
    \vspace{-0.3em}
    {\footnotesize  (a) 100 iterations after \quad \quad \quad \quad \quad \quad (b) 10,000 iterations after \\ \: Task 1 introduced  \:\:\: \quad \quad \quad \quad \quad \quad \quad  Task 1 introduced}
    \vspace{-0.5em}
    \caption{
        t-SNE visualization of 'bias-problem' in data distributions (class 0 to 3). (a) Only after 100 iterations of training after task 1 appears, learning is likely insufficient, and we can see that the features of new classes (class 2, 3) are biased towards the feature cluster of the existing class (\ie, class 1). (b) With more training iterations (10,000 iter), the features are well clustered by class.       
    } 
    \label{fig:feature_distrib}
    \vspace{-1em}
\end{figure}

Since novel classes continuously arrive in CL, both the data from previous tasks ($x_\text{old}$) and the current task ($x_\text{new}$) coexist.
While $\hat{f}(x_\text{old})$ are placed closer to their corresponding ETF classifier vectors $w_\text{new}$ by training, $\hat{f}(x_\text{new})$ are biased toward the cluster of $\hat{f}(x_\text{old})$, as we can see in Fig.~\ref{fig:feature_distrib}-(a), where $\hat{f}(x_\text{new})$ and $\hat{f}(x_\text{old})$ are the outputs of the model for inputs $x_\text{new}$ and $x_\text{old}$, respectively.
We call this `\emph{bias problem}.'
When $\hat{f}(x_\text{new})$ and $\hat{f}(x_\text{old})$ overlap due to bias and are optimized with the same objective function, training on the new class interferes with the representations of the old classes, dispersing the well-clustered $\hat{f}(x_\text{old})$~\citep{caccia2021new}. 
It destroys the ETF structure formed by $\hat{f}(x_\text{old})$, which hinders convergence towards neural collapse.
Although the bias problem exists even without a fixed ETF classifier~\cite{wu2019large, caccia2021new, chrysakis2023online}, it poses a greater problem when using a fixed ETF classifier, since the dispersed feature of old classes has to be restored to the corresponding fixed classifier vector.
In contrast, when using a learnable classifier, the dispersed representation of the old classes does not have to be restored to the original position, since the learnable classifier vector will be re-optimized from the dispersed feature. 
However, the angles between the classifier vector may become narrow, \ie, minority collapse occurs.

To accelerate convergence in the ETF structure, we propose to prevent novel classes from being biased towards existing classes when introduced, mitigating the bias problem.
Specifically, we train the model to avoid making predictions in favor of the existing classes for images that do not belong to them.
To this end, we propose to use preparatory data $x_p$ that is different from existing classes, obtained by transforming the samples in the episodic memory. 
By training with the preparatory data so that their representations differ from existing classes, we prevent biased predictions towards existing classes when a new class arrives.

For preparatory data to have a different representation from existing classes when trained, their image semantics should be distinguishable from existing classes, provided that they contain enough semantic information to be considered as an image (\ie, not noise).
There are various approaches to synthesize images with modified semantics, such as using generative models~\cite{chiaroni2020counter, graffieti2023generative, yang2023ganrec} or negative transformations~\cite{sinha2021negative, wang2022resmooth}.
We use negative transformations, as generative models are expensive in computation and storage~\cite{dedeoglu2023continual}, which is undesirable for CL scenarios with limited memory and computations.
On the contrary, negative transformations are relatively efficient in computation and storage~\cite{pankov2011learning} and is reported to create images with semantics different from the original image~\cite{wang2022resmooth}.
We empirically observed that data generated through negative transformations have a distinct semantic class from the original class while preserving its semantic information as an image (\ie, not as random noise), and we summarize the results in the supplementary material.


Formally, for a set of existing classes $Y$ and the set of possible transformations $G$, we randomly select $y\in Y$ and $g\in G$, and randomly retrieve a sample of class $y$ from memory and apply the transformation $g$ to obtain \emph{preparatory data} $x_p$.
We assign labels of the unseen classes to the preparatory data by a mapping function $m: Y\times G \rightarrow Y'$ where $Y'$ denotes a set of unseen classes, \ie, $Y' = \{y | y \notin Y,~ 1\leq y \leq K\}$ and $K$ is the total number of classifier vectors in $\mathbf{W}_\text{ETF} \in \mathbb{R}^{d \times K}$.
Thus, preparatory data $x_p$ from class $y$ and transformation $g$ is pulled towards the classifier vector $\mathbf{w}_{p}$, where $p = m(y, g)$.
When a new class $y_\text{new}$ is added to $Y$, we update $m$ by randomly assigning a new mapping $m(y_\text{new}, g)$ for $y_\text{new}$ and $g\in G$.

We compose the transformation set $G$ as a negative transformation that can modify semantic information (\ie, change the label), used in the self-supervised literature~\citep{gidaris2018unsupervised, feng2019self} and out-of-distribution (OOD) detection~\cite{sinha2021negative, wang2022resmooth, kim2022continual}.
Specifically, we use rotation by 90, 180, and 270 degrees ~\citep{feng2019self, gidaris2018unsupervised} as our negative transformation.
Since the vertical information of the image is more important than the horizontal information~\cite{jang2022pooling, tumpach2023temporal}, \eg, images of a car facing left or right are common, but a car flipped upside down is very rare compared to a car standing right, rotation by a large angle causes loss of semantic information of images~\cite{wang2022resmooth}.
Thus, we can use rotated data as semantically different images from images of existing classes, \ie, suitable for preparatory data. 
A more detailed analysis of various negative transformations is provided in the Supple.

We jointly train the model using real data (\ie, samples of existing classes) and preparatory data.
Formally, we write the objective for the model parameter $\theta$ as:
\begin{align}
    \argmin_\theta &\biggl[ \mathbb{E}_{(x,y)\sim \mathcal{M}}\mathcal{L}_\text{DR}\left(\hat{f}_\theta (x), y ; \mathbf{W}_{\text{ETF}}\right) \\
    +\lambda \, & \mathbb{E}_{(x',y')\sim \mathcal{M}, g\sim G} \mathcal{L}_\text{DR}\left(\hat{f}_\theta \left(g(x')\right), m(y', g) ; \mathbf{W}_{\text{ETF}}\right)\biggr],\nonumber    
\end{align}
where $\mathcal{M}$ denotes episodic memory, $(x, y)$ and $(x', y')$ denote the samples in $\mathcal{M}$, $G$ denotes the set of possible transformations, $g$ denotes a transformation in $G$, $m$ denotes the mapping function, and $\lambda$ is a hyperparameter for balancing real data and preparatory data. 


\begin{table*}[t!]
    \centering
    \resizebox{\textwidth}{!}{%
\begin{tabular}{lcccccccc}
\toprule
& \multicolumn{4}{c}{CIFAR-10} &
\multicolumn{4}{c}{CIFAR-100}\\ \cmidrule(lr){2-5} \cmidrule(lr){6-9}
&\multicolumn{2}{c}{Disjoint} &
\multicolumn{2}{c}{Gaussian-Scheduled} &
\multicolumn{2}{c}{Disjoint} &
\multicolumn{2}{c}{Gaussian-Scheduled} \\
Method & $A_\text{AUC} \ \uparrow$ & $A_{last} \ \uparrow$ & $A_\text{AUC} \ \uparrow$ & $A_{last} \ \uparrow$  & $A_\text{AUC} \ \uparrow$ & $A_{last} \ \uparrow$ & $A_\text{AUC} \ \uparrow$ & $A_{last} \ \uparrow$ \\ 
\cmidrule(lr){1-1} \cmidrule(lr){2-5} \cmidrule(lr){6-9}
EWC {\footnotesize \color{azure}{(Kirkpatrick \etal, 2017)}}
& 75.25$\pm$0.78 & 60.80$\pm$2.20 
& 59.62$\pm$0.31 & 64.24$\pm$1.97
& 52.08$\pm$0.83 & 41.55$\pm$0.85
& 38.22$\pm$0.50 & 42.52$\pm$0.58 \\
ER {\footnotesize \color{azure}{(Rolnick et al., 2019)}}
& 75.94$\pm$0.86 & 63.56$\pm$1.32
& 60.13$\pm$0.56 & 64.81$\pm$2.70
& 52.95$\pm$1.25 & 42.82$\pm$0.05
& 41.12$\pm$0.56 & 42.74$\pm$1.09 \\
ER-MIR {\footnotesize \color{azure}{(Aljundi \etal, 2019)}}
& 75.89$\pm$1.02 & 61.93$\pm$0.93
& 60.39$\pm$0.48 & 61.64$\pm$3.86 
& 52.93$\pm$1.44 & 42.47$\pm$0.13
& 41.19$\pm$0.63 & 42.93$\pm$1.18 \\
REMIND {\footnotesize \color{azure}{(Hayes \etal, 2020)}}
& 69.55$\pm$0.91 & 53.34$\pm$1.01
& 58.01$\pm$0.72 & 59.27$\pm$1.86
& 40.87$\pm$0.76 & 36.17$\pm$1.83
& 23.40$\pm$2.25 & 28.78$\pm$1.71 \\
DER++ {\footnotesize \color{azure}{(Buzzega \etal, 2020)}}
& 74.78$\pm$0.72 & 59.20$\pm$0.95
& 59.44$\pm$0.28 & 66.11$\pm$1.80 
& 38.16$\pm$1.57 & 38.55$\pm$2.11
& 29.38$\pm$2.58 & 38.20$\pm$3.13 \\
SCR {\footnotesize \color{azure}{(Mai \etal, 2021)}}
& 75.61$\pm$0.93 & 56.52$\pm$0.52
& 60.62$\pm$0.43 & 58.41$\pm$2.39
& 41.84$\pm$0.74 & 36.00$\pm$0.83
& 31.33$\pm$0.41 & 32.11$\pm$0.39 \\
ODDL {\footnotesize \color{azure}{(Ye \etal, 2022)}}
& 75.03$\pm$1.00 & 61.61$\pm$3.55
& 65.46$\pm$0.46 & 66.19$\pm$2.08
& 40.26$\pm$0.50 & 41.88$\pm$4.52
& 38.82$\pm$0.49 & 41.35$\pm$1.08 \\
MEMO {\footnotesize \color{azure}{(Zhou \etal, 2023)}}
& 73.21$\pm$0.49 & 62.47$\pm$3.38
& 59.26$\pm$0.90 & 62.01$\pm$1.17
& 40.60$\pm$1.11 & 39.87$\pm$0.46
& 23.41$\pm$1.63 & 32.74$\pm$2.11 \\
X-DER {\footnotesize \color{azure}{(Boschini \etal, 2023)}}
& 77.59$\pm$0.62 & 65.40$\pm$4.79
& 58.28$\pm$1.17 & 59.76$\pm$4.28
& 52.80$\pm$1.61 & 43.73$\pm$0.86 
& 41.94$\pm$0.57 & 46.00$\pm$1.04 \\
\cmidrule(lr){1-1} \cmidrule(lr){2-5} \cmidrule(lr){6-9}
EARL (Ours)
& \textbf{78.31$\pm$0.72} & \textbf{65.71$\pm$2.26}
& \textbf{69.52$\pm$0.19} & \textbf{70.41$\pm$1.97} 
& \textbf{57.12$\pm$1.22} & \textbf{44.40$\pm$0.68}
& \textbf{47.89$\pm$0.61} & \textbf{46.09$\pm$0.26} \\
\bottomrule
\\
\toprule
& \multicolumn{4}{c}{TinyImageNet} &
\multicolumn{4}{c}{ImageNet-200}\\ \cmidrule(lr){2-5} \cmidrule(lr){6-9}
&\multicolumn{2}{c}{Disjoint} &
\multicolumn{2}{c}{Gaussian-Scheduled} &
\multicolumn{2}{c}{Disjoint} &
\multicolumn{2}{c}{Gaussian-Scheduled} \\
Method & $A_\text{AUC} \ \uparrow$ & $A_{last} \ \uparrow$ & $A_\text{AUC} \ \uparrow$ & $A_{last} \ \uparrow$  & $A_\text{AUC} \ \uparrow$ & $A_{last} \ \uparrow$ & $A_\text{AUC} \ \uparrow$ & $A_{last} \ \uparrow$ \\ 
\cmidrule(lr){1-1} \cmidrule(lr){2-5} \cmidrule(lr){6-9}
EWC {\footnotesize \color{azure}{(Kirkpatrick \etal, 2017)}}
& 37.95$\pm$0.93 & 27.50$\pm$0.80
& 25.29$\pm$0.81 & 26.06$\pm$0.52
& 41.84$\pm$0.64 & 31.57$\pm$0.80
& 30.71$\pm$0.27 & 33.33$\pm$0.98 \\
ER {\footnotesize \color{azure}{(Rolnick \etal, 2019)}}
& 37.43$\pm$1.05 & 27.47$\pm$0.63
& 26.37$\pm$0.89 & 25.79$\pm$0.44
& 41.51$\pm$0.76 & 30.87$\pm$0.72
& 32.39$\pm$0.36 & 33.09$\pm$0.37 \\
ER-MIR {\footnotesize \color{azure}{(Aljundi \etal, 2019)}}
& 37.81$\pm$1.06 & 26.72$\pm$0.86
& 26.22$\pm$0.69 & 25.11$\pm$1.04
& 38.28$\pm$0.38 & 33.12$\pm$0.73
& 32.17$\pm$0.44 & 33.85$\pm$0.93 \\
REMIND {\footnotesize \color{azure}{(Hayes \etal, 2020)}}
& 28.37$\pm$0.13 & 27.68$\pm$0.45
& 10.19$\pm$0.60 & 14.90$\pm$1.49
& 39.25$\pm$0.93 & 31.98$\pm$0.84
& 30.23$\pm$0.62 & 33.98$\pm$0.09 \\
DER++ {\footnotesize \color{azure}{(Buzzega \etal, 2020)}}
& 39.38$\pm$0.60 & 29.36$\pm$0.81
& 27.23$\pm$1.94 & 31.53$\pm$0.80
& 43.50$\pm$0.31 & 34.56$\pm$0.50
& 35.22$\pm$0.26 & 38.38$\pm$0.97 \\
SCR {\footnotesize \color{azure}{(Mai \etal, 2021)}}
& 34.65$\pm$1.08 & 22.18$\pm$0.32
& 25.86$\pm$0.94 & 22.54$\pm$0.59
& 41.90$\pm$0.40 & 28.92$\pm$0.40
& 33.24$\pm$0.32 & 30.98$\pm$0.28 \\
MEMO {\footnotesize \color{azure}{(Zhou \etal, 2023)}}
& 27.36$\pm$0.61 & 27.57$\pm$0.52
& 10.82$\pm$1.23 & 18.03$\pm$1.36
& 41.55$\pm$0.23 & 34.19$\pm$1.47
& 32.54$\pm$0.39 & 36.11$\pm$1.06 \\
X-DER {\footnotesize \color{azure}{(Boschini \etal, 2023)}}
& 35.15$\pm$2.12 & 26.67$\pm$0.52
& 29.71$\pm$0.86 & 28.10$\pm$0.50
& 43.41$\pm$0.47 & 34.14$\pm$0.98
& 36.31$\pm$0.17 & 38.61$\pm$0.55 \\
\cmidrule(lr){1-1} \cmidrule(lr){2-5} \cmidrule(lr){6-9}
EARL (Ours)
& \textbf{41.77$\pm$1.26} & \textbf{29.65$\pm$0.20}
& \textbf{35.08$\pm$0.70} & \textbf{32.49$\pm$1.21} 
& \textbf{44.88$\pm$0.29} & \textbf{34.79$\pm$0.55}
& \textbf{39.14$\pm$0.47} & \textbf{38.83$\pm$0.35} \\
\bottomrule
\\
\end{tabular}}
\vspace{-1em}
\caption{Comparison of online CL methods on Disjoint and Gaussian Scheduled Setup for CIFAR-10, CIFAR-100, TinyImageNet and ImageNet-200.}
\vspace{-1.5em}
\label{tab:main}
\end{table*}

\subsection{Residual correction at inference}
\label{sec:residual}

Despite preparatory data training that accelerates convergence towards ETF, in online CL, new samples in a data stream hinder models from reaching the Terminal Phase of Training (TPT) and fully converging to the ETF during single-epoch training. 
When the output of the model $f(x)$ does not fully converge to the ETF classifier, the model would not perform well at all times, leading to poor anytime inference accuracy ($A_\text{AUC}$).

To address this issue, we want to correct the residual between $f(x)$ and the corresponding classifier vector $\mathbf{w}_{y}$ during inference, where $y$ is label of input $x$.

However, the discrepancy between the prediction on test data $x_\text{eval}$ and the ground-truth ETF classifier vector is not available in the inference phase.
In a similar situation, ResMem~\citep{yang2023resmem} stores the residual obtained from training data and uses them during inference, but they assume two-stage learning algorithms, \ie, standard training and inference rather than incremental learning and anytime inference.


Inspired by ResMem~\citep{yang2023resmem}, which uses residuals from the training in the inference stage, we propose to use the residual obtained from the training process to compensate for the remaining discrepancy between the classifiers and the features at inference.

To select which of the stored residuals to use during inference, we not only store the residual, but also $f(x)$ to choose the stored residual closest to $f(x_\text{infer})$. 
Therefore, we retain `feature-residual' pairs in a `feature-residual memory' $M = \{(\hat{h}_{i}, \mathbf{r}_{i}) \}_{i=1}^{N}$, where $\hat{h}_{i} = \hat{f}(x_i)$, $\mathbf{r}_i = \mathbf{w}_{y_i} - \hat{f}(x_i)$, where $N$ is the size of feature-residual memory, and $\mathbf{w}_{y_i}$ is the classifier vector for class $y_i$.

During inference, we select the $k$ nearest neighbor $\hat{h}$'s, \ie, $\{\hat{h}_{n_1}, \hat{h}_{n_1}, \dots, \hat{h}_{n_k}\}$ from $\{\hat{h}_i\}_{i=1}^{N}$, since using only the nearest residual for correction may lead to incorrect inference predictions if a wrong residual is selected from a different class.
Finally, following ResMem~\cite{yang2023resmem}, we calculate the residual correction term $\textbf{r}_\text{eval}$ by a weighted average of the corresponding residuals $\{\textbf{r}_{n_1}, \textbf{r}_{n_2}, \dots, \textbf{r}_{n_k}\}$, with weights $\{s_1, s_2, \dots, s_k\}$ that are inversely proportional to the distance from $\hat{f}(x_\text{eval})$, as:
\begin{equation}
    \mathbf{r}_\text{eval} = \sum_{i=1}^{k}s_i\mathbf{r}_i, \quad
    s_i = \frac{e^{-(\hat{f}(x_\text{eval}) - \hat{h_i})/\tau}}{\sum_{j=1}^{k} e^{-(\hat{f}(x_\text{eval}) - \hat{h}_j)/\tau}},
\end{equation}
where $\tau$ is a temperature hyperparameter.
We add the residual-correcting term $\mathbf{r}_\text{eval}$ to the model output $\hat{f}(x_\text{eval})$ to obtain the corrected output $\hat{f}(x_\text{eval})_\text{corrected}$ as:
\begin{equation}
     \hat{f}(x_\text{eval})_\text{corrected} = \hat{f}(x_\text{eval}) + \mathbf{r}_\text{eval}.
\end{equation}


\section{Experiments}
\label{sec:experiments}

\subsection{Experimental setup}
We perform experiments on four datasets: CIFAR-10, CIFAR-100, TinyImageNet, ImageNet-200, and ImageNet-1K. 
For all datasets, our experiments are conducted on both a disjoint setup~\citep{parisi2019continual} and a Gaussian scheduled setup~\citep{shanahan2021encoders, wang2022learning}. 
We report the average and standard deviation results in three different seeds, except ImageNet-1k due to computational resource constraints~\cite{bang2021rainbow, koh2021online}. 

To evaluate anytime inference performance, we use the area under the curve accuracy($A_{auc}$)~\citep{koh2021online, caccia2022anytime}, which measures the area under the accuracy curve. We also use last accuracy($A_{last}$) which measures accuracy at the end of training.
For detailed information about the experimental setup, refer to the Supplementary.

\vspace{-0.5em}
\paragraph{Baselines.}
We compare EARL with various CL methods in different categories: regularization (EWC~\citep{kirkpatrick2017overcoming}), network expansion (MEMO~\citep{zhou2022model}), replay (ER~\citep{rolnick2019experience}, ER-MIR~\citep{aljundi2019task}, REMIND~\citep{hayes2020remind}
, SCR~\citep{mai2021supervised}, ODDL~\citep{boschini2022class}), and distillation (DER++~\citep{buzzega2020dark}, X-DER~\cite{boschini2022class}).
For more details on the implementation of these methods and a comparison with NC-FSCIL~\citep{yang2023neural}, which attempts to induce neural collapse in the context of few-shot class incremental learning, please refer to the supplementary material.

\vspace{-0.5em}
\paragraph{Implementation details.}
We use three components to architect our model: a backbone network $g(\cdot)$, a projection MLP $p(\cdot)$, and a fixed ETF classifier $\mathbf{W}_\text{ETF}$ (\ie, our model $f$ can be defined as $f(x) = p \circ g(x)$).
For the projection layer, we attach an MLP projection layer $p_{\theta_p}$ to the output of the backbone network $g_{\theta_g}$, where $\theta_g$ and $\theta_p$ denote the parameters of the backbone network and the projection layer, respectively (\ie, the model $f$ can be defined as $f(x) = p \circ g(x)$), following ~\citep{chen2020simple, peng2022few, yang2023neural}. 
For all methods, we use ResNet-18~\citep{he2016deep} as the backbone network.

Following~\citep{koh2021online, ye2022task}, we employ memory-only training, where a random batch is selected from memory at each iteration. 
Furthermore, for episodic memory sampling, EARL uses the Greedy Balanced Sampling strategy~\citep{prabhu2020gdumb}. 
We describe the details about hyperparameters and the pseudocode of EARL in Supplementary for the sake of space.

\subsection{Results}

\begin{table}[t!]
    \centering
    \resizebox{0.48\textwidth}{!}{%
        \begin{tabular}{lcccc}
        \toprule
        \multirow{3}{*}{Method} & \multicolumn{4}{c}{ImageNet-1K} \\ \cmidrule(lr){2-5}
        & \multicolumn{2}{c}{Disjoint} & \multicolumn{2}{c}{Gaussian-Scheduled} \\
        & $A_\text{AUC} \ \uparrow$ & $A_{last} \ \uparrow$ & $A_\text{AUC} \ \uparrow$ & $A_{last} \ \uparrow$ \\ 
        \cmidrule(lr){1-1} \cmidrule(lr){2-3} \cmidrule(lr){4-5}
        EWC {\footnotesize \color{azure}{(Kirkpatrick \etal, 2017)}}
        & 30.17  & 18.78
        & 17.09  & 20.15 \\
        ER {\footnotesize \color{azure}{(Rolnick \etal, 2019)}}
        & 30.18  & 18.96
        & 17.17  & 18.39 \\
        ER-MIR {\footnotesize \color{azure}{(Aljundi \etal, 2019)}}
        & 31.68  & 19.87
        & 19.37  & 16.30 \\
        REMIND {\footnotesize \color{azure}{(Hayes \etal, 2020)}}
        & 32.47  & \textbf{24.59}
        & 17.42  & 19.79 \\
        DER++ {\footnotesize \color{azure}{(Buzzega \etal, 2020)}}
        & 31.08  & 18.98
        & 23.73  & 23.14 \\
        SCR {\footnotesize \color{azure}{(Mai \etal, 2021)}}
        & 26.72  & 13.60
        & 19.82  & 15.84 \\
        MEMO {\footnotesize \color{azure}{(Zhou \etal, 2023)}}
        & 30.45  & 19.08
        & 14.06  & 20.12 \\
        X-DER {\footnotesize \color{azure}{(Boschini \etal, 2023)}}
        & 31.67  & 18.18
        & 25.18  & 12.51 \\
        \cmidrule(lr){1-1} \cmidrule(lr){2-5}
        EARL (Ours)
        & \textbf{34.33} & 23.19
        & \textbf{30.53} & \textbf{29.48} \\
        \bottomrule
        \end{tabular}}
    \vspace{-1em}
    \caption{Comparison of online CL methods on Disjoint and Gaussian Scheduled Setup for ImageNet-1K.}
    \vspace{-1.5em}
    \label{tab:imagenet}
\end{table}

We first compare the accuracy of online continual learning methods, including EARL, and summarize the results in Table~\ref{tab:main}.
As shown in the table, EARL outperforms other baselines on all benchmarks, both in disjoint and Gaussian-scheduled setups, except $A_\text{last}$ in ImageNet-1K disjoint setup. 
In particular, high $A_\text{AUC}$ suggests that EARL outperforms other methods for all the time that the data stream is provided to the model, which implies that it can be used for inference at anytime.

Furthermore, EARL, SCR, ER, MIR, DER, and ODDL do not use task boundary information during training, \ie, \emph{task-free}, in contrast to EWC, X-DER, MEMO, and REMIND which use task-boundary information, \ie, \emph{task-aware}.
Nevertheless, EARL outperforms these methods even in the disjoint setup where utilizing task boundary information is advantageous due to abrupt distribution shifts at the task boundaries, except MEMO in ImageNet-1K disjoint.
Since MEMO not only uses the task boundary information but also expands the network per task, this extra advantage of using a larger model is emphasized in large-scale datasets such as ImageNet which require a large learning capacity of the model. 
Please refer to the supplementary material for details of the computational budget comparison among the baseline methods.

\vspace{-1.2em}
\paragraph{Memory budget.}
We consider not only the size of episodic memory, but also the size of the model, logits, and other components, following MEMO~\citep{zhou2022model}.
In the case of the memory budget for CIFAR-10 and the architecture of ResNet-18 (size of 42.6MB), EARL incurs an additional memory cost of 0.2MB to store feature-residual pairs. EWC requires an additional cost of 85.2MB due to the need to store Fisher Information per parameter and the previous model. MEMO, which expands the network per task, incurs an additional cost of 32MB per task. DER also requires an additional cost of 0.1MB to store logits during distillation. 


\subsection{Ablation study}
\label{sec:ablations}
We conduct an ablation study on the two components of EARL, preparatory data training and residual correction, and summarize the results in Table \ref{tab:ablation}. 
Although both components contribute to performance improvement, preparatory data training shows a larger gain in performance.
Training with preparatory data improves the baseline by 2.7\% $\sim$ 3.3\% in $A_{auc}$ and 2.1\% $\sim$ 4.4\% in $A_{last}$. 
When combined with residual correction, we observe further gains across all metrics in both disjoint and Gaussian-scheduled setups.

\begin{figure}[h!]
  \centering
  \vspace{-1.5em}
  \begin{center}
    \includegraphics[width=0.8\linewidth]{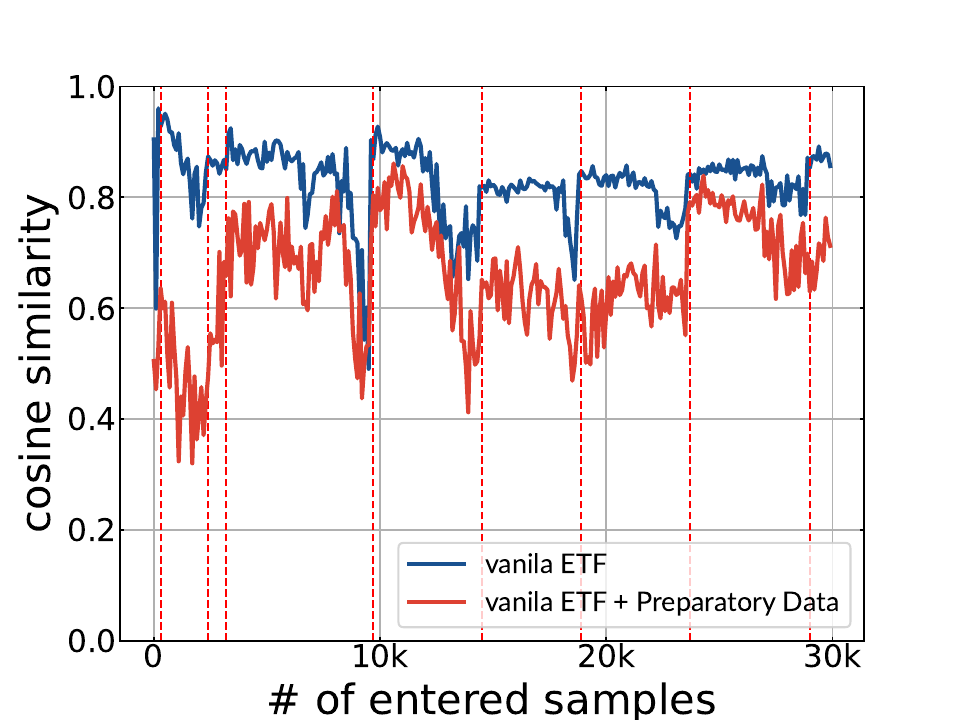}
  \end{center}
  \vspace{-1em}
  \caption{Average similarity between features of the most recently added class's samples and the closest classifier vectors of the old classes (CIFAR-10, Gaussian Scheduled). Baseline is a vanilla ETF model trained only using episodic memory.}
  \label{fig:bias_solve}
\end{figure}

\begin{table*}[h!]
    \vspace{-1em}
    \centering
    \resizebox{\textwidth}{!}{
    \begin{tabular}{lcccccccc}
    \toprule
    \multirow{3}{*}{\makecell[c]{METHOD}} & \multicolumn{4}{c}{CIFAR-10} & \multicolumn{4}{c}{CIFAR-100} \\
        \cmidrule(lr){2-5} \cmidrule(lr){6-9}
        & \multicolumn{2}{c}{Disjoint} & \multicolumn{2}{c}{Gaussian-Scheduled} &  \multicolumn{2}{c}{Disjoint} & \multicolumn{2}{c}{Gaussian-Scheduled} \\ 
         & $A_\text{AUC} \ \uparrow$ & $A_\text{last} \ \uparrow$ & $A_\text{AUC} \ \uparrow$ & $A_\text{last} \ \uparrow$ & $A_\text{AUC} \ \uparrow$ & $A_\text{last} \ \uparrow$ & $A_\text{AUC} \ \uparrow$ & $A_\text{last} \ \uparrow$  \\
         \cmidrule(lr){1-1} \cmidrule(lr){2-3} \cmidrule(lr){4-5} \cmidrule(lr){6-7} \cmidrule(lr){8-9}
    
    EARL (Ours)
    & \textbf{78.61$\pm$0.72} & \textbf{66.01$\pm$2.26}
    & \textbf{69.62$\pm$0.19} & \textbf{70.91$\pm$1.97}
    & \textbf{57.42$\pm$1.24} & \textbf{44.60$\pm$0.65}
    & \textbf{48.19$\pm$0.61} & \textbf{46.10$\pm$0.26} \\ 
    
    (-) RC 
    & 77.78$\pm$0.52 & 65.52$\pm$1.63
    & 68.37$\pm$0.10 & 70.17$\pm$1.08
    & 56.28$\pm$1.40 & 44.29$\pm$1.04
    & 47.05$\pm$0.71 & 46.07$\pm$0.52 \\ 
    
    (-) RC \& PDT 
    & 74.77$\pm$0.77 & 61.10$\pm$3.12
    & 65.60$\pm$0.25 & 67.39$\pm$0.78
    & 52.91$\pm$1.05 & 41.08$\pm$0.60
    & 44.34$\pm$0.55 & 44.45$\pm$0.48 \\ 
    \bottomrule
    \end{tabular}
    }
    \vspace{-.5em}
    \caption{Ablation Study. RC and PDT refer to preparatory data training (Sec.~\ref{sec:preparatory}) and the residual correction (Sec.~\ref{sec:residual}), respectively.}
    \label{tab:ablation}
    \vspace{-1em}
\end{table*}

\begin{figure*}[h!]
    \centering
    \includegraphics[width=0.93\linewidth]{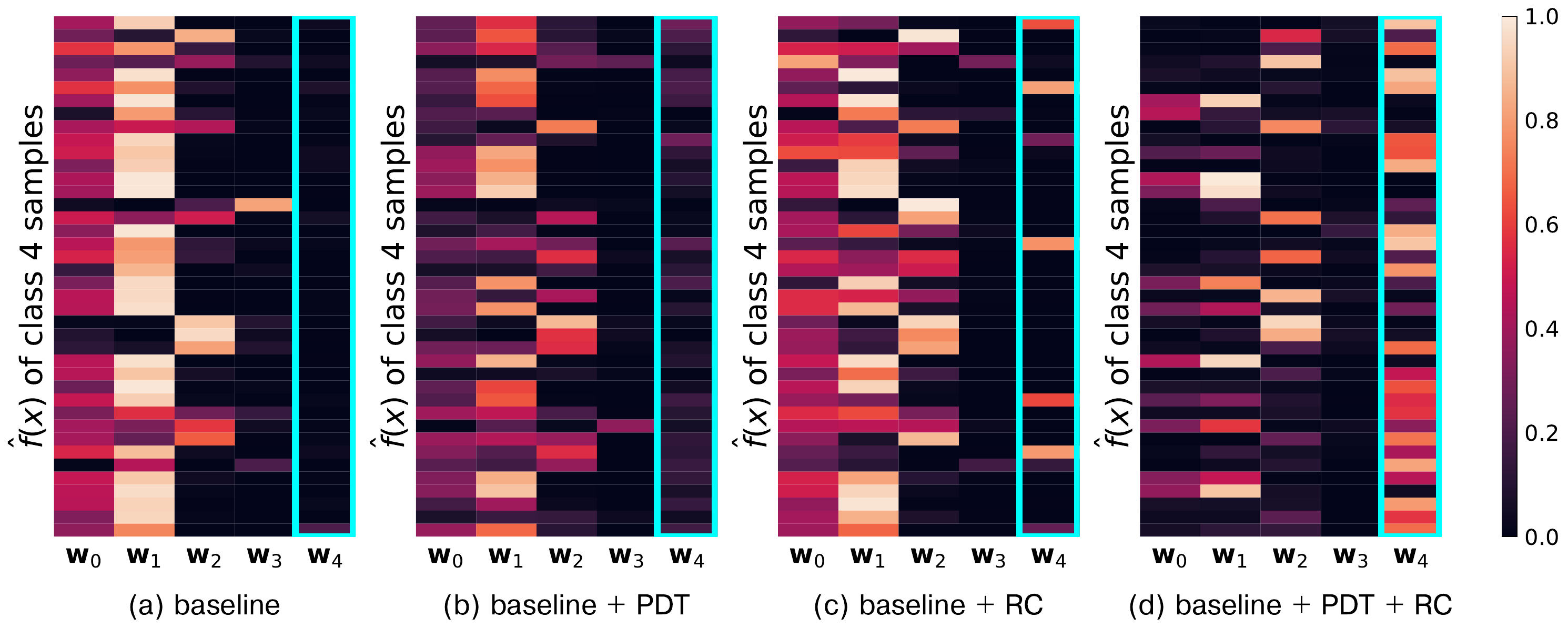}
    \vspace{-0.75em}
    \caption{
        Cosine similarity between features $\hat{f}(x)$ for class 4 and the ETF classifier vectors $\mathbf{w}_i$ at the 50$^{th}$ iteration after the introduction of class 4 in the Gaussian Scheduled CIFAR-10 setup. 
        As we can see in the cyan highlighted box, EARL promotes the convergence of $\hat{f}(x)$ for class 4 toward the ground truth classifier vector $\mathbf{w}_4$.
    } 
    \label{fig:comp_baselines}
    \vspace{-1em}
\end{figure*}

For further analysis, Fig.~\ref{fig:comp_baselines} visualizes the cosine similarity between the output features of 50 randomly selected test set samples of novel class `4' and the classifier vectors $\mathbf{w}_i$.
In baseline (a), the features of the class 4 are strongly biased towards the classifier vectors of the old classes (\ie, $\mathbf{w}_0$, $\mathbf{w}_1$, $\mathbf{w}_2$, $\mathbf{w}_3$), rather than the correct classifier, $\mathbf{w}_4$. (c) When residual correction is used, some samples show high similarity with $\mathbf{w}_4$ compared to the baseline.

However, since incorrect residuals can be added due to the bias problem, more samples have high similarity with $\mathbf{w}_2$ than the baseline (\ie, wrong residuals are added). 
(b) When using preparatory data training, the bias toward $\mathbf{w}_0$ and $\mathbf{w}_1$ classes significantly decreases compared to the baseline. 
Fig.~\ref{fig:bias_solve} also shows the effect of preparatory data, which reduces the similarity between the novel class features and existing classes.
(d) Using both residual correction and preparatory data training shows a remarkable alignment with the ground truth classifier $\mathbf{w}_4$.
A more detailed analysis of the ablation results is provided in the Supple.

\section{Conclusion}
To better learn online data in a continuous data stream without multiple epoch training, we propose to induce neural collapse, which aligns last layer activations to the corresponding classifier vectors in the representation space.
Unlike in offline CL, it is challenging to induce neural collapse in online CL due to insufficient training epochs and continuously streamed new data.
We first observe that the bias of the new class towards existing classes slows the convergence of features toward neural collapse.

To mitigate the bias, we propose synthesizing preparatory data for unseen classes by transforming the samples of existing classes.
Using the preparatory data for training, we accelerate neural collapse in an online CL scenario.
Additionally, we propose residual correction to resolve the remaining discrepancy toward neural collapse at inference, which arises due to the continuous stream of new data.
In our empirical evaluations, the proposed methods outperform state-of-the-art online CL methods in various datasets and setups, especially with high performance on anytime inference.

\vspace{-1em}
\paragraph{Limitations and Future Work.}
Since our work uses ETF structure, it has an inherent limitation that the number of possible classifier vectors in the ETF classifier is limited by the dimension of the embedding space.
Considering lifelong learning, where the number of new classes goes to infinity, it is interesting to explore the idea of dynamically expanding the ETF structure so that the model can continually learn the ever-increasing number of concepts in the real world.

\footnotesize{\vspace{-1em}
\paragraph{Acknowledgement.} This work was partly supported by the NRF grant (No.2022R1A2C4002300, 20\%) and IITP grants (No.2022-0-00077 (10\%), No.2022-0-00113 (25\%), No.2022-0-00959 (10\%), No.2022-0-00871 (10\%), No.2020-0-01361 (5\%, Yonsei AI), No.2021-0-01343 (5\%, SNU AI), No.2021-0-02068 (5\%, AI Innov. Hub), No.2022-0-00951 (5\%)) funded by the Korea government (MSIT) and CARAI grant funded by DAPA and ADD (UD230017TD) 5\%. This work was done while J. Choi was with Yonsei University and partially while H. Lee, M. Seo and H. Kim was with LG AI Research.
}

{
    \small
    \bibliographystyle{ieeenat_fullname}
    \bibliography{main}
}

\clearpage
\noindent\textbf{Note:} {\color{blue}Blue} characters denote the reference of the main paper. \blfootnote{\hspace{-2em}$^\dagger$: Work done while interning at LG AI Research. \\ $~~^*$: Indicates corresponding authors.}

\section{Anytime Inference in Online CL (\outref{L164})}
\label{sec:anytime_infer}
In online CL, new data continuously arrive in a stream rather than in a large chunk (\eg, task unit). 
Several previous works [\outref{2}]\citep{kim2021continual} train the model only after a large chunk of new data accumulates, which leads to poor inference performance on new data during data accumulation, since the model is not trained during the accumulation period [\outref{6, 27}]. 

However, inference queries can occur at any time including the data accumulation, \ie, \emph{anytime inference}, whose importance has been emphasized by recent research~\citep{doan2022continual, banerjee2023verse} [\outref{17, 27, 36}]. 
Consequently, we focus not only on $A_{last}$, which is measured after the learning has finished for all data, but also on $A_\text{AUC}$, which measures the average accuracy during training, \ie, `anytime inference' performance [\outref{27}].

\section{Analysis about Negative Transformation (\outref{L332})}
\label{sec:hard_rotation}

\begin{table}[h!]
    \vspace{-1em}
    \centering
    \resizebox{0.95\linewidth}{!}{
        \begin{tabular}{ccccc}
            \toprule
            \multirow{2}{*}{Negative Transformation}
            & \multicolumn{2}{c}{Gaussian-Scheduled} & \multicolumn{2}{c}{Disjoint}  \\ 
            & $A_\text{AUC} \ \uparrow$ & $A_\text{last} \ \uparrow$ & $A_\text{AUC} \ \uparrow$ & $A_\text{last} \ \uparrow$ \\ 
            \cmidrule(lr){1-1} \cmidrule(lr){2-3} \cmidrule(lr){4-5} 

            Patch Permutation
            & 66.37$\pm$0.36 & 69.04$\pm$1.39
            & 75.34$\pm$0.88 & 63.73$\pm$2.05 \\ 
            
            Negative Cutmix
            & 65.26$\pm$0.22 & 64.48$\pm$1.07
            & 72.86$\pm$0.13 & 56.56$\pm$3.25 \\ 
            
            Gaussian Noise
            & 64.83$\pm$0.23 & 66.97$\pm$1.40
            & 74.18$\pm$0.22 & 63.32$\pm$1.52 \\ 
            
            Negative Rotation
            & \textbf{69.52$\pm$0.13} & \textbf{70.28$\pm$1.77}
            & \textbf{77.86$\pm$0.71} & \textbf{69.50$\pm$1.94} \\ 
            
            \bottomrule
            \\
        \end{tabular}}

    \vspace{-1em}
    \caption{Comparison of various negative transformations to obtain preparatory data in CIFAR-10} 
    \vspace{-1em}
    \label{tab:transform_comp}
\end{table}
Negative transformation is a transformation that modifies semantic information of original images and is widely used in self-supervised learning [\outref{4, 5}] and out-of-distribution (OOD) detection [\outref{25, 47, 50}].
We consider various negative transformations such as negative rotation (\ie, rotation by 90, 180, and 270 degrees) [\outref{15, 18}], patch permutation~\cite{doersch2015unsupervised}, negative CutMix, and Gaussian noise.
Negative Cutmix is a kind of Cutmix~\cite{yun2019cutmix} that finely divides an image into small patches and mixes it with another finely divided image.
Gaussian noise refers to filling all the image pixels with a Gaussian random noise. 
Even though Gaussian noise is not a kind of transformation, it is a straightforward approach to generate synthetic input that can be distinguished from the existing images; thus, we also consider this.

Among them, we choose to use rotations 90, 180, and 270 degrees as the negative transform set $T$, because the rotation transformation is simple to implement, widely applicable from low- to high-resolution images, and also outperforms other transformations in our empirical evaluations, as we can see in Tab.~\ref{tab:transform_comp}. 
Unlike rotation transformation, which preserves image continuity while modifying semantic information, patch permutation or Negative Cutmix could cause discontinuities at the boundary of patches~\cite{kimata2022objectmix, lee2022sagemix}, leading to loss of image features. 

\begin{figure*}[t!]
    \centering
    \includegraphics[width=0.9\linewidth]{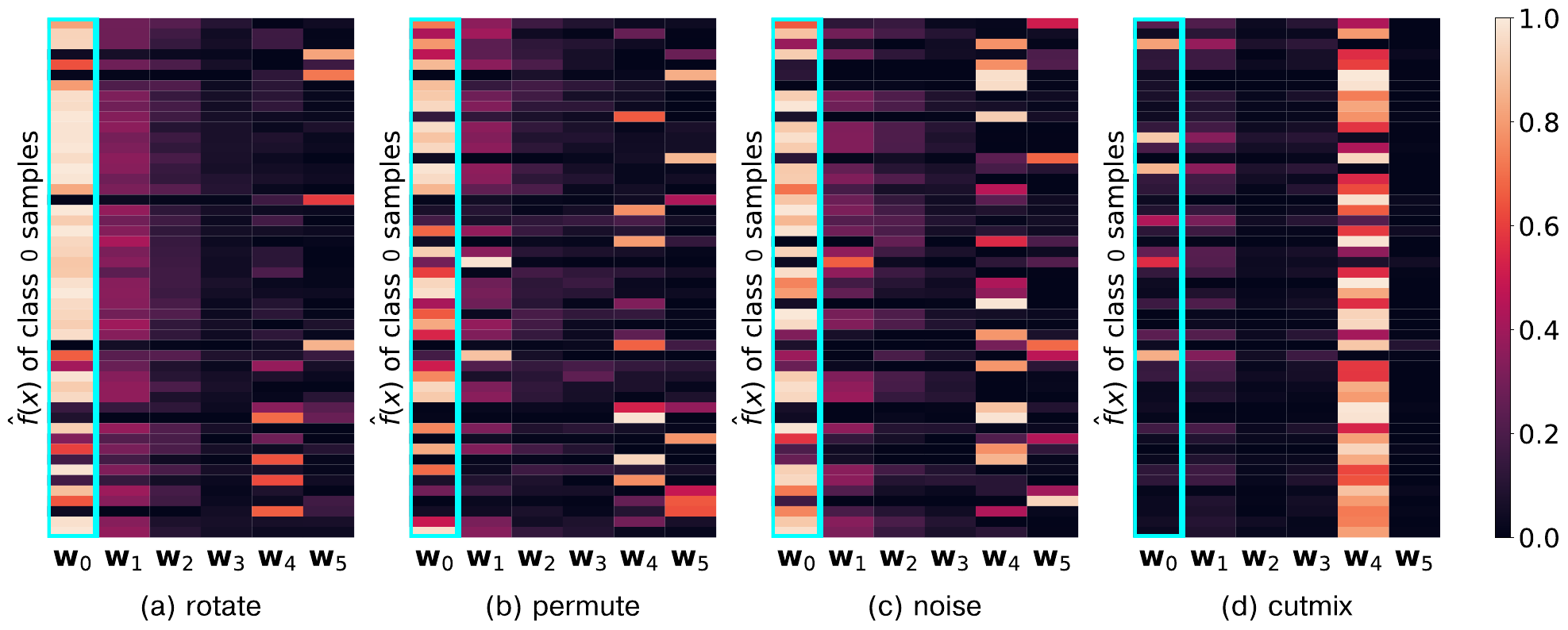}
    \vspace{-.5em}
    \caption{
        Cosine similarity between the features $\hat{f}(x)$ for class 0 and the ETF classifier vectors $\mathbf{w}_i$ at the 10000$^{th}$ iteration after the introduction of class 0 in the Gaussian Scheduled CIFAR-10 setup. 
    } 
    \label{fig:comp_transforms}
\end{figure*}

Furthermore, to compare the effect of various negative transformations, we compare the cosine similarities between the features and the corresponding ground truth classifier as shown in Fig.~\ref{fig:comp_transforms}.
As we can see in the cyan highlighting box, the rotation transformation promotes the convergence of $\hat{f}(x)$ for class $0$ towards the ground truth classifier vector $\mathbf{w}_0$.

\section{Effect of Negative Rotation Transformation (\outref{L304})}
\label{sec:negative_transform}
We use negative rotation, which rotates images by 90, 180, and 270 degrees [\outref{15, 18}].
Since the vertical information of the image is more important than the horizontal information [\outref{22, 49}] as mentioned in \outref{Sec.4.2}, rotation by a large angle changes the semantic information [\outref{50}].
To empirically verify that the negative rotation transformation alters the semantic information of the image, we generate four datasets by rotating the original data by 0, 90, 180, and 270 degrees and train the model using each dataset. 
Subsequently, we performed inference on all four datasets to check whether they have different semantics. 

Specifically, we used CIFAR-10-R90, CIFAR-10-R180, and CIFAR-10-R270, created by applying rotation transformations of 90, 180, and 270 degrees to CIFAR-10 dataset, respectively.
We summarize the results in Tab.~\ref{tab:transform_effect}.

\begin{table}[h!]
    \vspace{-.5em}
    \centering
    \resizebox{0.95\linewidth}{!}{
        \begin{tabular}{ccccc}
            \toprule
            \multirow{2}{*}{Train Dataset} & \multicolumn{4}{c}{Evaluation Dataset}\\ 
            & CIFAR-10 & CIFAR-10-R90 & CIFAR-10-R180 & CIFAR-10-R270 \\
            \cmidrule(lr){1-1} \cmidrule(lr){2-5}

            CIFAR-10
            & \textbf{92.9} & 31.7 & 34.7 & 31.8 \\ 

            CIFAR-10-R90
            & 32.1 & \textbf{92.6} & 31.9 & 35.3 \\ 

            CIFAR-10-R180
            & 33.7 & 30.9 & \textbf{92.5} & 32.3 \\ 

            CIFAR-10-R270
            & 32.4 & 34.3 & 31.4 & \textbf{92.4} \\ 
            
            \bottomrule
            \\
        \end{tabular}
        }

    \vspace{-1em}
    \caption{Comparison of the inference accuracy of the rotated datasets after training with each rotated dataset.}
    \vspace{-1em}
    \label{tab:transform_effect}
\end{table}

\section{Prevent Forgetting with ETF Classifier}
We compare the performance and forgetting between a fixed ETF classifier and a learnable classifier. We summarize the results in Tab.~\ref{tab:etf_forgetting}. 
The fixed ETF classifier not only effectively prevents forgetting, but also has high accuracy.

\begin{table}[h!]
    \vspace{-0.5em}
    \centering
    \resizebox{\linewidth}{!}{
        \begin{tabular}{ccccc}
            \toprule
            \multirow{2}{*}{Classifier}
            & \multicolumn{2}{c}{CIFAR-10} & \multicolumn{2}{c}{CIFAR-100} \\ 
            & $A_\text{AUC} \ \uparrow$ & Forgetting$\ \downarrow$ & $A_\text{AUC} \ \uparrow$ & Forgetting$\ \downarrow$ \\
            \cmidrule(lr){1-1} \cmidrule(lr){2-3} \cmidrule(lr){4-5}
            Learnable
            & 66.02$\pm$0.18 & 6.54$\pm$1.02 & 44.37$\pm$0.84 & 9.04$\pm$1.45 \\ 
            Fixed ETF
            & \textbf{69.61$\pm$0.35} & \textbf{3.75$\pm$2.21} & \textbf{47.78$\pm$0.69} & \textbf{7.41$\pm$1.02} \\ 
            \bottomrule
            \\
        \end{tabular}
        }
    \vspace{-1.5em}
    \caption{Comparison of Forgetting and $A_\text{AUC}$ between a learnable classifier and a classifier with fixed ETF structure on CIFAR-10/100.}
    \label{tab:etf_forgetting}
\end{table}

\section{Comparison of Baselines with Computational Constraint}
\label{sec:comp_const}
\begin{figure*}[h!]
    \centering
    \includegraphics[width=0.9\linewidth]{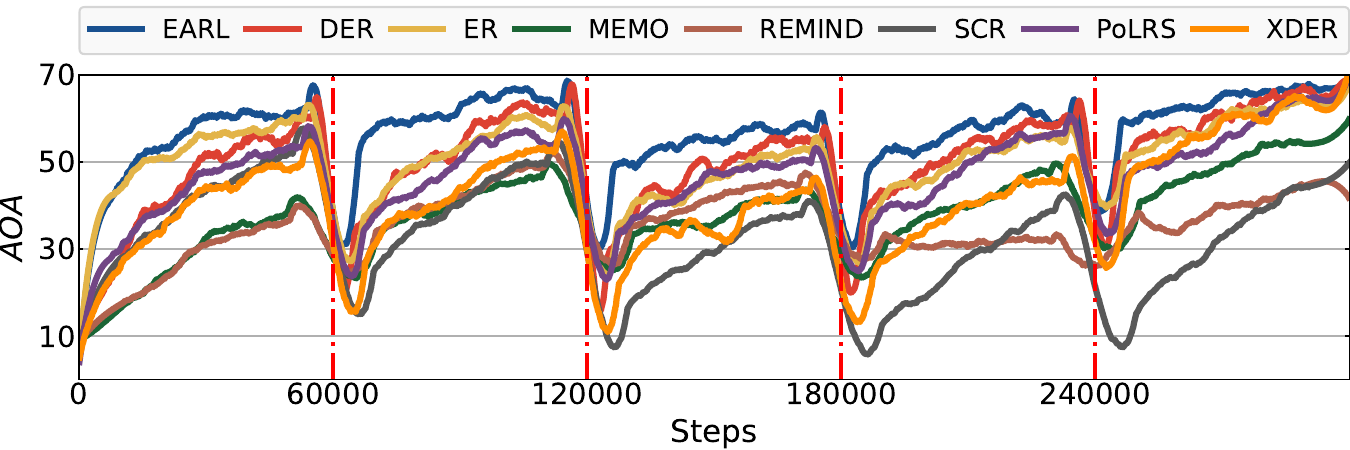}
    \vspace{-0.5em}
    \caption{Online accuracy in TinyImageNet Disjoint. Red lines denote task boundary}
    \label{fig:tiny_online}
    \vspace{-.5em}
\end{figure*}

Following ~\cite{seo2023budgeted}[\outref{17}], we measure the number of FLOPs required for each method to complete one iteration. We then calculate the relative FLOPs with respect to the ER [\outref{43}], which is the simplest method, and adjust the total training iteration to align the total FLOPs used for the entire training process. 
With this computational constraint, we compare online CL methods on disjoint and Gaussian scheduled setup for CIFAR-10, CIFAR-100, TinyImageNet and ImageNet-200. 
Furthermore, we measure the Average Online Accuracy ($AOA$)[\outref{7, 17}], which uses the newly encountered streaming data for evaluation before incorporating it into the training process.
We plot the online accuracy on TinyImageNet in Fig.~\ref{fig:tiny_online}.

\begin{table*}[t!]
    \centering
    \resizebox{\textwidth}{!}{%
\begin{tabular}{lcccccccccccc}
\toprule
& \multicolumn{6}{c}{CIFAR-10} & \multicolumn{6}{c}{CIFAR-100} \\ 
\cmidrule(lr){2-7} \cmidrule(lr){8-13}
&\multicolumn{3}{c}{Disjoint} &
\multicolumn{3}{c}{Gaussian-Scheduled} &
\multicolumn{3}{c}{Disjoint} &
\multicolumn{3}{c}{Gaussian-Scheduled} \\
Method & $A_\text{AUC} \ \uparrow$ & $A_{last} \ \uparrow$ & $AOA \ \uparrow$ & $A_\text{AUC} \ \uparrow$ & $A_{last} \ \uparrow$ & $AOA \ \uparrow$ & $A_\text{AUC} \ \uparrow$ & $A_{last} \ \uparrow$  & $AOA \ \uparrow$  & $A_\text{AUC} \ \uparrow$ & $A_{last} \ \uparrow$  & $AOA \ \uparrow$  \\ 
\cmidrule(lr){1-1} \cmidrule(lr){2-7} \cmidrule(lr){8-13}
EWC {\footnotesize \color{azure}{(Kirkpatrick \etal, 2017)}}
& 77.30$\pm$0.91 & 62.78$\pm$1.25 & 88.70$\pm$0.89
& 60.91$\pm$0.30 & 63.56$\pm$3.04 & 83.22$\pm$0.16
& 52.94$\pm$1.00 & 43.27$\pm$0.41 & 62.04$\pm$0.52
& 40.89$\pm$0.40 & 42.80$\pm$0.89 & 55.29$\pm$0.18 \\
ER {\footnotesize \color{azure}{(Rolnick et al., 2019)}}
& 76.89$\pm$0.96 & 63.92$\pm$2.36 & 88.47$\pm$0.78
& 60.71$\pm$0.10 & 66.71$\pm$2.49 & 82.47$\pm$0.17
& 53.13$\pm$1.28 & 41.97$\pm$0.21 & 62.06$\pm$0.73
& 41.30$\pm$0.22 & 44.10$\pm$0.14 & 55.28$\pm$0.33 \\
ER-MIR {\footnotesize \color{azure}{(Aljundi \etal, 2019)}}
& 75.08$\pm$0.10 & 63.13$\pm$3.32 & 87.27$\pm$1.52
& 57.25$\pm$0.71 & 59.34$\pm$2.12 & 79.29$\pm$0.69
& 50.17$\pm$0.91 & 42.17$\pm$0.50 & 57.74$\pm$0.31
& 35.52$\pm$0.57 & 41.29$\pm$0.71 & 46.44$\pm$0.55 \\
REMIND {\footnotesize \color{azure}{(Hayes \etal, 2020)}}
& 67.73$\pm$0.42 & 49.33$\pm$1.30 & 81.30$\pm$1.17
& 55.94$\pm$1.06 & 51.73$\pm$4.11 & 77.41$\pm$1.27
& 40.76$\pm$0.46 & 37.64$\pm$1.13 & 48.18$\pm$0.98
& 22.55$\pm$1.58 & 30.27$\pm$1.94 & 33.02$\pm$1.19 \\
DER++ {\footnotesize \color{azure}{(Buzzega \etal, 2020)}}
& 75.71$\pm$1.46 & 58.49$\pm$3.16 & \textbf{88.93$\pm$1.00}
& 59.14$\pm$1.15 & 66.16$\pm$3.83 & 82.14$\pm$0.76
& 41.39$\pm$0.80 & 42.03$\pm$0.81 & 57.00$\pm$0.30
& 28.63$\pm$1.93 & 37.10$\pm$2.23 & 47.39$\pm$1.38 \\
SCR {\footnotesize \color{azure}{(Mai \etal, 2021)}}
& 75.89$\pm$0.47 & 57.90$\pm$2.45 & 87.91$\pm$1.82
& 60.38$\pm$0.25 & 63.02$\pm$3.71 & 78.73$\pm$0.48
& 36.74$\pm$1.72 & 30.54$\pm$0.99 & 34.56$\pm$1.26
& 26.53$\pm$1.00 & 27.59$\pm$0.96 & 29.88$\pm$0.93 \\
ODDL {\footnotesize \color{azure}{(Ye \etal, 2022)}}
& 75.91$\pm$0.87 & 61.89$\pm$4.47 & 88.61$\pm$1.50
& 61.02$\pm$0.60 & 65.25$\pm$1.01 & 83.58$\pm$0.69
& 53.16$\pm$1.18 & 42.89$\pm$0.64 & 61.19$\pm$0.78
& 42.46$\pm$0.30 & 44.26$\pm$1.26 & 56.33$\pm$0.36 \\
MEMO {\footnotesize \color{azure}{(Zhou \etal, 2023)}}
& 72.59$\pm$0.18 & 63.29$\pm$5.07 & 86.33$\pm$1.81
& 57.88$\pm$1.30 & 61.67$\pm$2.32 & 79.04$\pm$0.82
& 39.48$\pm$0.73 & 37.53$\pm$0.64 & 50.06$\pm$0.13
& 22.46$\pm$1.40 & 33.27$\pm$2.39 & 36.02$\pm$1.21 \\
X-DER {\footnotesize \color{azure}{(Boschini \etal, 2023)}}
& 75.99$\pm$1.22 & 65.69$\pm$4.05 & 83.28$\pm$2.05
& 57.73$\pm$0.76 & 66.59$\pm$2.45 & 76.63$\pm$0.32
& 47.22$\pm$1.21 & 43.72$\pm$0.56 & 50.73$\pm$0.78
& 36.81$\pm$0.56 & 46.20$\pm$0.53 & 47.59$\pm$0.33 \\
\cmidrule(lr){1-1} \cmidrule(lr){2-7} \cmidrule(lr){8-13}
EARL (Ours)
& \textbf{78.54$\pm$0.48} & \textbf{66.16$\pm$0.84} & 88.10$\pm$0.58
& \textbf{69.77$\pm$0.05} & \textbf{71.46$\pm$1.84} & \textbf{86.00$\pm$0.17}
& \textbf{57.65$\pm$1.30} & \textbf{45.15$\pm$0.05} & \textbf{67.09$\pm$0.69}
& \textbf{48.05$\pm$0.67} & \textbf{47.27$\pm$0.69} & \textbf{63.69$\pm$0.76} \\
\bottomrule
\\
\toprule
& \multicolumn{6}{c}{TinyImageNet} & \multicolumn{6}{c}{ImageNet-200} \\ 
\cmidrule(lr){2-7} \cmidrule(lr){8-13}
&\multicolumn{3}{c}{Disjoint} &
\multicolumn{3}{c}{Gaussian-Scheduled} &
\multicolumn{3}{c}{Disjoint} &
\multicolumn{3}{c}{Gaussian-Scheduled} \\
Method & $A_\text{AUC} \ \uparrow$ & $A_{last} \ \uparrow$ & $AOA \ \uparrow$ & $A_\text{AUC} \ \uparrow$ & $A_{last} \ \uparrow$ & $AOA \ \uparrow$ & $A_\text{AUC} \ \uparrow$ & $A_{last} \ \uparrow$  & $AOA \ \uparrow$  & $A_\text{AUC} \ \uparrow$ & $A_{last} \ \uparrow$  & $AOA \ \uparrow$  \\ 
\cmidrule(lr){1-1} \cmidrule(lr){2-7} \cmidrule(lr){8-13}
EWC {\footnotesize \color{azure}{(Kirkpatrick \etal, 2017)}}
& 37.57$\pm$0.83 & 27.43$\pm$0.89 & 49.58$\pm$0.42
& 26.35$\pm$0.84 & 25.61$\pm$0.23 & 37.77$\pm$0.25
& 41.85$\pm$0.64 & 31.57$\pm$0.76 & 64.10$\pm$0.26
& 32.22$\pm$0.27 & 32.71$\pm$0.94 & 54.68$\pm$0.31 \\
ER {\footnotesize \color{azure}{(Rolnick et al., 2019)}}
& 37.29$\pm$0.81 & 27.04$\pm$0.40 & 49.66$\pm$0.39
& 26.37$\pm$0.90 & 25.92$\pm$0.38 & 37.66$\pm$0.50
& 41.65$\pm$0.62 & 32.18$\pm$0.22 & 64.22$\pm$0.23
& 32.43$\pm$0.51 & 32.85$\pm$0.27 & 54.72$\pm$0.50 \\
ER-MIR {\footnotesize \color{azure}{(Aljundi \etal, 2019)}}
& 37.73$\pm$0.84 & 27.40$\pm$0.22 & 48.05$\pm$0.15
& 24.00$\pm$0.73 & 25.10$\pm$0.41 & 32.34$\pm$0.46
& 39.88$\pm$0.44 & 33.27$\pm$0.66 & 59.87$\pm$0.14
& 28.65$\pm$0.26 & 33.30$\pm$1.84 & 48.36$\pm$0.44 \\
REMIND {\footnotesize \color{azure}{(Hayes \etal, 2020)}}
& 29.05$\pm$0.83 & 27.30$\pm$0.59 & 36.43$\pm$0.83
& 10.22$\pm$0.93 & 16.59$\pm$0.70 & 16.85$\pm$1.17
& 38.32$\pm$0.73 & 31.60$\pm$0.46 & 39.80$\pm$1.85
& 30.46$\pm$0.37 & 33.36$\pm$0.82 & 38.13$\pm$1.17 \\
DER++ {\footnotesize \color{azure}{(Buzzega \etal, 2020)}}
& 39.24$\pm$0.80 & 29.77$\pm$1.16 & 49.30$\pm$0.81
& 27.27$\pm$2.13 & 31.40$\pm$0.89 & 40.19$\pm$1.47
& 44.67$\pm$0.17 & 32.65$\pm$0.54 & 65.74$\pm$0.21
& 37.11$\pm$0.28 & 37.85$\pm$0.74 & 59.81$\pm$0.33 \\
SCR {\footnotesize \color{azure}{(Mai \etal, 2021)}}
& 34.07$\pm$1.16 & 24.24$\pm$0.43 & 33.28$\pm$0.93
& 25.62$\pm$0.90 & 24.28$\pm$0.24 & 29.36$\pm$0.41
& 41.94$\pm$0.31 & 28.53$\pm$0.37 & 61.97$\pm$0.40
& 33.27$\pm$0.46 & 31.41$\pm$0.31 & 54.12$\pm$0.48 \\
MEMO {\footnotesize \color{azure}{(Zhou \etal, 2023)}}
& 27.84$\pm$0.38 & 27.52$\pm$0.52 & 37.35$\pm$0.26
& 9.45$\pm$0.60 & 17.30$\pm$0.65 & 17.50$\pm$0.49
& 38.68$\pm$0.39 & 31.70$\pm$0.53 & 58.92$\pm$0.53
& 32.09$\pm$0.26 & 35.95$\pm$1.40 & 50.88$\pm$0.10 \\
X-DER {\footnotesize \color{azure}{(Boschini \etal, 2023)}}
& 35.68$\pm$0.38 & 27.15$\pm$1.66 & 41.36$\pm$1.41
& 23.51$\pm$2.43 & 24.80$\pm$4.06 & 32.17$\pm$3.71
& 44.03$\pm$0.37 & 33.58$\pm$1.17 & 56.40$\pm$0.08
& 34.92$\pm$0.24 & 38.29$\pm$1.12 & 53.75$\pm$0.23 \\
\cmidrule(lr){1-1} \cmidrule(lr){2-7} \cmidrule(lr){8-13}
EARL (Ours)
& \textbf{42.19$\pm$1.18} & \textbf{29.50$\pm$0.78} & \textbf{56.41$\pm$0.08}
& \textbf{34.79$\pm$0.67} & \textbf{31.68$\pm$0.21} & \textbf{49.77$\pm$0.13}
& \textbf{45.01$\pm$0.36} & \textbf{34.25$\pm$0.69} & \textbf{67.08$\pm$0.34}
& \textbf{39.04$\pm$0.22} & \textbf{38.95$\pm$0.50} & \textbf{60.80$\pm$0.13} \\
\bottomrule
\\
\end{tabular}}
\vspace{-1em}
\caption{Comparison of online CL methods on Disjoint and Gaussian Scheduled Setup for CIFAR-10, CIFAR-100, TinyImageNet and ImageNet-200 with computational constraint.}
\label{tab:main_const}
\end{table*}

\section{Comparison of Baselines on CLEAR-10/100}
To evaluate the setups in which the domain of classes changes over time, we compare CL methods with the CLEAR benchmark and summarize the result in Tab.~\ref{tab:clear}. 
In our experiments, we address a more realistic scenario, called the domain-class-IL setup, where both novel classes and domain shifts take place.
This contrasts with domain-IL setups, where all classes are initially provided and only the data distribution changes over time.

EARL significantly outperforms the baselines in CLEAR-10/100 benchmarks and achieves high performance.
We perform experiments under the computational constraints mentioned in Sec.~\ref{sec:comp_const}.

\begin{table*}[h!]
    \centering
    \resizebox{0.7\linewidth}{!}{
        \begin{tabular}{ccccccc}
            \toprule
            \multirow{2}{*}{Method}
            & \multicolumn{3}{c}{CLEAR10} & \multicolumn{3}{c}{CLEAR100} \\
            & $A_\text{AUC} \ \uparrow$ & $A_\text{last} \ \uparrow$ & $AOA \ \uparrow$ & $A_\text{AUC} \ \uparrow$ & $A_\text{last} \ \uparrow$ & $AOA \ \uparrow$ \\
            \cmidrule(lr){1-1} \cmidrule(lr){2-4} \cmidrule(lr){5-7}
            EWC
            & 70.88$\pm$1.15 & 69.46$\pm$2.40 & 76.96$\pm$1.16 & 45.74$\pm$0.40 & 47.61$\pm$0.54 & 46.15$\pm$0.28 \\
            ER
            & 70.70$\pm$1.22 & 68.86$\pm$3.00 & 76.65$\pm$1.12 &  45.59$\pm$0.91 & 47.89$\pm$1.11 & 46.05$\pm$0.18  \\
            ER-MIR
            & 68.21$\pm$0.94 & 65.58$\pm$2.33 & 74.53$\pm$1.10 &  43.21$\pm$1.03 & 46.60$\pm$1.21 & 42.24$\pm$0.47 \\
            REMIND
            & 66.48$\pm$1.93 & 66.91$\pm$1.19 & 72.34$\pm$1.78 &  36.67$\pm$0.76 & 47.90$\pm$0.58 & 35.49$\pm$0.09 \\
            DER++
            & 71.93$\pm$0.90 & 70.41$\pm$2.67 & 77.51$\pm$1.47 &   47.34$\pm$0.63 & 49.63$\pm$0.75 & 47.02$\pm$0.33 \\
            SCR++
            & 73.32$\pm$0.85 & 70.81$\pm$1.69 & 77.90$\pm$1.11 &  44.67$\pm$0.77 & 44.70$\pm$0.72 & 46.15$\pm$0.28 \\
            MEMO
            & 65.04$\pm$1.72 & 62.64$\pm$2.67 & 71.94$\pm$1.80 &  44.35$\pm$0.54 & 46.47$\pm$1.58 & 41.24$\pm$0.12 \\
            PoLRS
            & 65.65$\pm$1.88  & 61.06$\pm$6.17 & 71.67$\pm$1.53 &  
            41.17$\pm$1.83  & 41.62$\pm$2.56 & 40.15$\pm$1.50 \\
            X-DER
            & 69.77$\pm$0.85 & 68.67$\pm$3.04 & 74.76$\pm$1.23 & 44.76$\pm$1.21 & 49.01$\pm$1.31 & 43.60$\pm$0.65 \\
            \cmidrule(lr){1-1} \cmidrule(lr){2-4} \cmidrule(lr){5-7}
            EARL
            & \textbf{77.85$\pm$0.96} & \textbf{76.51$\pm$1.97} & \textbf{81.51$\pm$1.17} & \textbf{56.97$\pm$0.25} & \textbf{59.03$\pm$1.09} & \textbf{55.35$\pm$0.06} \\
            \bottomrule
            \\
        \end{tabular}
        }
    \vspace{-1.5em}
    \caption{Comparison of online CL methods on CLEAR-10/100 with computational constraint.}
    \vspace{-1.1em}
    \label{tab:clear}
\end{table*}

\section{Comparison of Computational Budget (\outref{L})}
EARL requires an additional computation for residual correction, which calculates $k$ nearest features, and FLOPs of additional cost can be formulated as $d \times N \times 3 + k \times N$, where $d$ is the dimension of stored features, $N$ is the total number of feature-residual pairs, and $k$ denotes $\text{top}_k$ in $k$-NN. 
The first term is for calculating the distances between the stored features and the feature of the inference image and is multiplied by 3 since it involves subtraction, squaring, and addition operations.
The second term is the cost of selecting $\text{top}_k$ features among $N$ features.
Compared to the cost incurred by the naive inference process, which requires forward flops of the model, it involves a very minimal amount of additional cost. Taking the example of ImageNet-200 with the ResNet-18 architecture that has $\frac{1}{3}$ GFLOPs in the model forward, only {0.5}\% of additional cost is consumed, since we use $N=2,000(=10\times200)$ and $k=15$.

\section{Details About Experiment Setup (\outref{L404})}\
\label{sec:exp_detail}
This paper focuses on online class-incremental learning in two types of setups: disjoint [\outref{35}] and Gaussian scheduled~\cite{koh2023online} [\outref{45, 51}]. 
In the disjoint setup, each class is assigned to a specific task, \ie, tasks do not share any classes.
On the other hand, in the Gaussian scheduled setup, the arrival time of each class follows a Gaussian distribution $\mathcal{N}(\mu_i, \sigma)$. 
Since the class distribution is shifted every time step, the Gaussian scheduled setup is a boundary-free setup.
We set $\sigma$ to 0.1 and $\mu_i$, mean of the class $i$, to $\frac{i}{N}$, where $N$ is the number of classes. 

\section{Effect of $k$}
\begin{figure}[h!]
    \vspace{-0.5em}
    \centering
    \includegraphics[width=\linewidth]{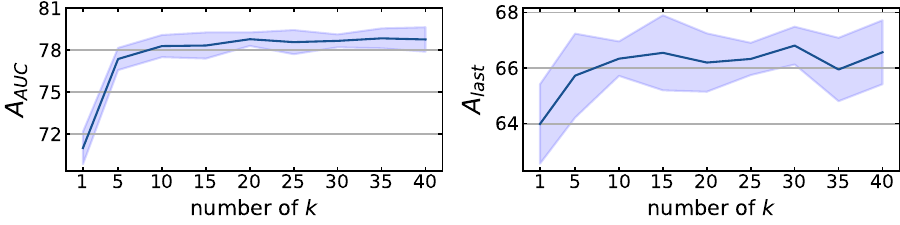}
    \caption{Performance variations of $A_\text{AUC}$ and $A_{last}$ with respect to the change in $k$ in the CIFAR-10 disjoint setup.}
    \label{fig:k_effect}
    \vspace{-0.5em}
\end{figure}

$k$, which is a hyperparameter that determines the number of the nearest features from the inference image used to weight-sum to calculate the residual, has a negligible impact, except for $k=1$ as incorrect residual largely affects predictions. We illustrate the consistency of accuracy at various values of $k$ in Fig.~\ref{fig:k_effect}.

\section{Hyperparameters (\outref{L428})}
\label{sec:appx:hyperparam}
For all methods, we use Adam optimizer~\cite{kingma15adam} with a constant learning rate (LR) of 0.0003.
For data augmentation, we use RandAugment~\cite{cubuk2020randaugment}. 
For hyperparameters such as iteration, memory size, and number of tasks for each dataset, we follow prior works [\outref{2, 27, 40}].
Specifically, we set the number of iterations as 1, 3, 3, 0.25, and 0.25, and memory sizes as 500, 2,000, 4,000, 4,000, and 10000 for the CIFAR-10, CIFAR-100, TinyImageNet, ImageNet-200, and ImageNet-1k datasets, respectively. 
To ensure a fair comparison among methods, we use the same batch size in all methods.
Specifically, we use 16, 32, 32, 256, and 256 for CIFAR-10, CIFAR-100, TinyImageNet, ImageNet-200, and ImageNet-1k datasets, respectively.
Note that the number of preparatory data is included in the batch size for EARL, \ie, batch size = the number of retrieved samples from memory + the number of preparatory data.

To ensure that EARL does not depend on a specific dataset, we use the following hyperparameters for all datasets: $\tau=0.9, k=15, d=4096$ and $\lambda=1$, where $\tau$ refers to the temperature parameter during residual correction, $k$ refers to the number of samples for $\text{top}_k$ in $k$-NN during residual correction, $d$ refers to the output dimension of the projection layer, and $\lambda$ refers to the loss balancing parameter between real data training and preparatory data training. 
In addition, for $N$, the total number of stored feature-residual pairs in EARL, we used $N=10*\left| C\right|$ for all dataset, where $\left|C\right|$ is the number of classes seen so far.

\section{Implementation of Baselines (\outref{L587})}
\label{sec:implementation}

Some of the baselines assumed multi-epoch training, \ie, offline CL, so we modified them to be used in online CL for comparisons with our proposed method and other baselines. 

\paragraph{REMIND.}
Remind is a feature-replay method and freezes front layers after base initialization, offline training phase using a subset of the data during the initial stage of training.
We modified the base initialization of REMIND [\outref{20}] to be suitable for use in online CL.
In offline CL setup, REMIND freezes 7 layers after base initialization.
However, as shown in Table~\ref{tab:layer_comp}, in online CL, it is not suitable to freeze many layers compared to offline CL, as the model is not sufficiently trained. 
We studied various numbers of frozen layers and revealed that freezing 6 layers (\ie, 3 blocks) is the optimal number for freezing in ResNet-18 in CIFAR-10 and CIFAR-100. 
Compared to freezing 7 layers, which is the original proposed freezing criterion,  REMIND performed comparatively well when freezing 6 layers in the online CL, despite the decrease in the number of stored features due to the larger sizes of stored features.
Regarding the hyperparameters used in REMIND, such as the size of the codebook and the number of codebooks for product quantization, we performed additional hyperparameter search experiments with CIFAR-100, as shown in Table~\ref{tab:codebooknum_comp}, and used the same hyperparameter in all remaining datasets.

For CIFAR-10, CIFAR-100, and TinyImageNet, which have a relatively small number of images, we increased the proportion of the initialization sample base from 10$\%$ to 60$\%$ of the total samples, since 10$\%$ images are not enough for the lower level layers to represent highly transferable features, leading to low performance.
In ImageNet-200 and ImageNet-1k, we followed the original setting (\ie, 10$\%$ base initialization samples and freezing 7 layers after base initialization), following [\outref{20}].

\begin{table*}[h!]
    \resizebox{1.0\linewidth}{!}{
        \begin{tabular}{ccccccccc}
            \toprule
            \multirow{3}{*}{Frozen Layers} & \multicolumn{4}{c}{CIFAR-10} & \multicolumn{4}{c}{CIFAR-100} \\
            \cmidrule(lr){2-5} \cmidrule(lr){6-9}
            & \multicolumn{2}{c}{Disjoint} & \multicolumn{2}{c}{Gaussian-Scheduled} &  \multicolumn{2}{c}{Disjoint} & \multicolumn{2}{c}{Gaussian-Scheduled}  \\ 
            & $A_\text{AUC} \ \uparrow$ & $A_\text{last} \ \uparrow$ & $A_\text{AUC} \ \uparrow$ & $A_\text{last} \ \uparrow$ & $A_\text{AUC} \ \uparrow$ & $A_\text{last} \ \uparrow$ & $A_\text{AUC} \ \uparrow$ & $A_\text{last} \ \uparrow$ \\
             
            \cmidrule(lr){1-1} \cmidrule(lr){2-3} \cmidrule(lr){4-5} \cmidrule(lr){6-7} \cmidrule(lr){8-9}

4 Layers
& 66.63$\pm$1.24 & 46.24$\pm$0.12
& 54.55$\pm$0.71 & 45.56$\pm$1.54
& 38.73$\pm$0.27 & 30.19$\pm$0.66
& 23.76$\pm$0.86 & 26.94$\pm$0.37 \\ 

5 Layers
& \textbf{70.11$\pm$0.66} & 51.01$\pm$0.79
& 56.47$\pm$0.96 & 52.00$\pm$1.94
& 41.87$\pm$0.05 & \textbf{38.81$\pm$0.41}
& 24.79$\pm$1.73 & 32.12$\pm$2.81 \\ 

6 Layers
& 69.55$\pm$0.91 & 47.28$\pm$3.92
& \textbf{57.15$\pm$0.71} & \textbf{53.40$\pm$0.70}
& \textbf{41.92$\pm$0.06} & 37.64$\pm$1.09
& \textbf{25.56$\pm$1.10} & \textbf{34.08$\pm$1.02} \\ 

7 Layers
& 68.59$\pm$0.10 & \textbf{53.71$\pm$2.27}
& 55.37$\pm$0.60 & 52.53$\pm$1.98
& 40.52$\pm$0.17 & 37.42$\pm$0.81
& 23.94$\pm$1.30 & 33.28$\pm$1.43 \\ 

\bottomrule
\\
\end{tabular}
}
\vspace{-1em}
\caption{REMIND performance as a function of the number of frozen layers in ResNet-18 with CIFAR-10 and CIFAR-100. Rather than freezing all the layers except the last layer (\ie, freezing 7 Layers), continuously updating the last two layers and fixing the rest (\ie, freezing 6 Layers) shows the best performance due to the limitation of online CL.} 
\vspace{-0.5em}
\label{tab:layer_comp}
\end{table*}

\begin{table*}[h!]
    \resizebox{1.0\linewidth}{!}{
        \begin{tabular}{ccccccccc}
            \toprule
            \multirow{3}{*}{$\sharp$ of Codebooks} & \multicolumn{4}{c}{CIFAR-100} \\
            \cmidrule(lr){2-5} \cmidrule(lr){6-9}
            & \multicolumn{2}{c}{Disjoint} & \multicolumn{2}{c}{Gaussian-Scheduled} \\ 
            & $A_\text{AUC} \ \uparrow$ & $A_\text{last} \ \uparrow$ & $A_\text{AUC} \ \uparrow$ & $A_\text{last} \ \uparrow$ \\
            \cmidrule(lr){1-1} \cmidrule(lr){2-3} \cmidrule(lr){4-5}

8
& 41.07$\pm$0.27 & 36.97$\pm$0.27
& 25.17$\pm$1.02 & \textbf{34.14$\pm$1.18} \\

16
& 41.84$\pm$0.29 & 36.48$\pm$0.34
& 25.37$\pm$0.99 & 33.30$\pm$0.26 \\ 

32
& \textbf{41.92$\pm$0.06} & \textbf{37.64$\pm$1.09}
& \textbf{25.56$\pm$1.10} & 34.08$\pm$1.02 \\ 

64
& 40.13$\pm$0.15 & 33.50$\pm$0.46
& 24.39$\pm$0.85 & 29.48$\pm$0.04 \\ 

\bottomrule
\\
\end{tabular}

{
\begin{tabular}{ccccccccc}
    \toprule
    \multirow{3}{*}{Codebook Size} & \multicolumn{4}{c}{CIFAR-100} \\
    \cmidrule(lr){2-5}
    & \multicolumn{2}{c}{Disjoint} & \multicolumn{2}{c}{Gaussian-Scheduled}  \\ 
    & $A_\text{AUC} \ \uparrow$ & $A_\text{last} \ \uparrow$ & $A_\text{AUC} \ \uparrow$ & $A_\text{last} \ \uparrow$ \\
     
    \cmidrule(lr){1-1} \cmidrule(lr){2-3} \cmidrule(lr){4-5}

256
& \textbf{41.92$\pm$0.06} &\textbf{ 37.64$\pm$1.09}
& 25.56$\pm$1.10 & \textbf{34.08$\pm$1.02} \\ 

512
& 41.48$\pm$0.19 & 36.98$\pm$0.51
& \textbf{25.85$\pm$1.39} & 33.01$\pm$1.99 \\ 

1024
& 41.28$\pm$0.18 & 36.48$\pm$0.34
& 25.13$\pm$0.90 & 31.83$\pm$0.19 \\ 

2048
& 41.11$\pm$0.21 & 36.10$\pm$0.33
& 25.05$\pm$0.90 & 31.33$\pm$0.11 \\ 

\bottomrule
\\
\end{tabular}
}
}
\vspace{-1em}
\caption{REMIND performance as a function of different codebook sizes and number of codebooks with CIFAR-100. Original hyperparameters (codebook size: 256, number of codebooks: 32) consistently show the best performance in the online CL setting. The same hyperparameters were used uniformly for all datasets.} 
\vspace{-0.5em}
\label{tab:codebooknum_comp}
\end{table*}

\paragraph{MEMO.}
MEMO [\outref{63}] retrieves samples that are relatively close to the class mean feature at every task boundary and stores them in episodic memory for replay. 
However, in online CL, the model can not access all data for the current task, \ie, it continuously updates the memory using stream data from the current task. 
Therefore, the sampling strategy of MEMO is replaced by class-balanced random sampling, which is used to replace the sampling strategy of RM [\outref{2}] for online CL in [\outref{27}].


\section{Comparison between NC-FSCIL and Vanilla ETF}
NC-FSCIL [\outref{56}] is a recently proposed offline CL method that attempts to induce neural collapse in few-shot class incremental learning (FSCIL), a CL setup with few training samples per class. 
NC-FSCIL uses the fixed ETF classifier, a backbone network $f$, and a projection layer. 
It freezes the backbone network after training the base task and further fine-tunes the projection layer in the following incremental tasks by using the replay memory $\mathcal{M}^{(t)}$, which stores the class-wise mean features $\mathbf{h}_c$ retrieved from the backbone network for each old class $c$ as follows:
\begin{equation}
    \mathcal{M}^{(t)} = \left\{\mathbf{h}_c|c\in\bigcup_{j=0}^{t-1}C^{(j)}\right\}, ~~ 1\le t\le T, 
\end{equation}
where $\mathbf{h}_c = \text{Avg}_i\left\{f\left(x_i, \theta\right)|y_i=c\right\}$, $T$ refers to the total number of tasks, and $C^{(j)}$ refers to the set of classes for task $j$.


Considering that our setup is neither an offline CL nor a few-shot class incremental learning setup, for a fair comparison with online CL methods, we modify the freezing strategy of NC-FSCIL.
Instead of freezing the whole backbone, we freeze only 6 layers, which achieved the best performance in online CL as shown in Table \ref{tab:layer_comp}.

Despite the high performance of NC-FSCIL in offline CL, it has a lower performance than that of vanilla ETF (\ie, baseline of EARL without preparatory data training and residual correction) in online CL, as shown in Tab.~\ref{tab:ncfscil}.

\begin{table*}[h!]
    \resizebox{1.0\linewidth}{!}{
        \begin{tabular}{ccccccccc}
            \toprule
            \multirow{3}{*}{Methods} & \multicolumn{4}{c}{CIFAR-10} & \multicolumn{4}{c}{CIFAR-100} \\
            \cmidrule(lr){2-5} \cmidrule(lr){6-9}
            & \multicolumn{2}{c}{Disjoint} & \multicolumn{2}{c}{Gaussian-Scheduled} &  \multicolumn{2}{c}{Disjoint} & \multicolumn{2}{c}{Gaussian-Scheduled}  \\ 
            & $A_\text{AUC} \ \uparrow$ & $A_\text{last} \ \uparrow$ & $A_\text{AUC} \ \uparrow$ & $A_\text{last} \ \uparrow$ & $A_\text{AUC} \ \uparrow$ & $A_\text{last} \ \uparrow$ & $A_\text{AUC} \ \uparrow$ & $A_\text{last} \ \uparrow$ \\
            \cmidrule(lr){1-1} \cmidrule(lr){2-3} \cmidrule(lr){4-5} \cmidrule(lr){6-7} \cmidrule(lr){8-9}

NC-FSCIL
& 68.09$\pm$0.69 & 48.05$\pm$1.69
& 53.12$\pm$0.77 & 46.11$\pm$0.36
& 39.54$\pm$0.79 & 34.83$\pm$1.27
& 30.91$\pm$1.46 & 35.73$\pm$1.13 \\ 

Vanilla ETF
& \textbf{75.27$\pm$0.77} & \textbf{62.10$\pm$4.12}
& \textbf{65.80$\pm$0.25} & \textbf{67.79$\pm$0.78}
& \textbf{52.91$\pm$1.05} & \textbf{41.08$\pm$0.60}
& \textbf{33.34$\pm$0.55} & \textbf{44.45$\pm$0.48} \\ 
\bottomrule
\\
\end{tabular}
}
\vspace{-1em}
\caption{Comparison between NC-FSCIL and Vanilla ETF (EARL w/o Preparatory data training and Residual correction) on CIFAR-10 and CIFAR-100.} 
\label{tab:ncfscil}
\end{table*}

\section{Pseudocode for the Our Method (\outref{L428})}
\label{sec:pseudocode}

Algorithm~\ref{algo:algo_train} and Algorithm~\ref{algo:sdp} provide detailed pseudocode for EARL.

\begin{algorithm*}[t!]
\caption{Training Phase}
\label{algo:algo_train}
\begin{algorithmic}[1]
\State \textbf{Input} model $f_\theta$, Memory $\mathcal{M}$,  Residual Memory $\mathcal{M}_{\text{RES}}$, Training data stream $\mathcal{D}$, ETF classifier $\mathbf{W}$, Negative transformation $\mathcal{R}_r$, Learning rate $\mu$

\For{$(x, y) \in \mathcal{D}$}{\small\color{azure}\Comment{Sample arrives from training data stream D}}
    \State \textbf{Update} $\mathcal{M} \leftarrow \text{ClassBalancedSampler}\left(\mathcal{M}, (x, y)\right)${\small\color{azure}\Comment{Update memory}}
    \State \textbf{Sample} $(X, Y) \leftarrow \text{RandomRetrieval}(\mathcal{M}) ${\small\color{azure}\Comment{Get batch $(X, Y)$ from memory}}
    \State \textbf{Sample} $(X', Y') \leftarrow \text{RandomRetrieval}(\mathcal{M}) ~~ ${\small\color{azure}\Comment{Get batch $(X', Y')$ to make preparatory data}}
    \State $(X_{\text{p}}, Y_{\text{p}}) \leftarrow \mathcal{R}_r(X',Y') ${\small\color{azure}\Comment{Negative transformation for preparatory data training}}
    \State $\hat{f_\theta}(X)=\frac{f_\theta(X)}{|f_\theta(X)|}, \hat{f_\theta}(X_p)=\frac{f_\theta(X_p)}{|f_\theta(X_p)|}${\small\color{azure}\Comment{Normalize model output}}
    \State $\mathbf{r}=\mathbf{W}_Y - \hat{f}_\theta(X)${\small\color{azure}\Comment{Calculate Residuals}}
    \State \textbf{Update} $\mathcal{M}_{\text{RES}} \leftarrow \left(\hat{f}_\theta(X),\mathbf{r}\right)${\small\color{azure}\Comment{Update feature-residual memory}}
    \State $\mathcal{L}(X, Y, X_p, Y_p; \theta, \mathbf{W})=L_{DR}(\hat{f_\theta}(X), \mathbf{W}_Y) + L_{DR}(\hat{f_\theta}(X_p), \mathbf{W}_{Y_p})${\small\color{azure}\Comment{Calculate dot-regression loss}}
    \State \textbf{Update} $\theta \leftarrow \theta - \mu\cdot\nabla_\theta \mathcal{L}(X, Y, X_p, Y_p; \theta, \mathbf{W})${\small\color{azure}\Comment{Update model}}

\EndFor
\State \textbf{Output} $f_{\theta}$
\end{algorithmic}
\end{algorithm*}

\begin{algorithm*}[t!]
\caption{Inference Phase}
\label{algo:sdp}
\begin{algorithmic}[1]
\State \textbf{Input} model $f_\theta$, inference input $x_\text{eval}$, Residual Memory $\mathcal{M}_{\text{RES}}$, ETF classifier $\mathbf{W}$, number of nearest neighbors $k$, softmax temperature $\tau$


\State $\{(\hat{h}_i, \mathbf{r}_i)\}_{i=1}^{\vert\mathcal{M}_{\text{RES}}\vert} \leftarrow \mathcal{M}_{\text{RES}} ${\small\color{azure}\Comment{Get residual and features from residual memory}}
\State $\hat{f_\theta}(x_\text{eval})=\frac{f_\theta(x_\text{eval})}{|f_\theta(x_\text{eval})|}${\small\color{azure}\Comment{Normalize model output}}
\State $\{n_1, n_2, ..., n_k\} \leftarrow k\text{-}\argmin_i \left(\left|\hat{f}_\theta(x_\text{eval})-\hat{h}_i\right|\right)${\small\color{azure}\Comment{Calculate $k$ nearest neighbor features}}
\State $s_{n_i} = \frac{e^{-(\hat{f}(x_\text{eval}) - \hat{h}_{n_i})/\tau}}{\sum_{j=1}^{k} e^{-(\hat{f}(x_\text{eval}) - \hat{h}_{n_j})/\tau}}${\small\color{azure}\Comment{calculate residual weights}}
\State $\mathbf{r} = \sum_{i=1}^{k}s_{n_i}\mathbf{r}_{n_i}${\small\color{azure}\Comment{Calculate residual-correcting term}}
\State $\hat{f}_\theta(x_\text{eval})_\text{corrected} \leftarrow \hat{f}_\theta(x_\text{eval})+\mathbf{r}${\small\color{azure}\Comment{Add residual on features}}
\State $y_\text{pred}=\argmax_y(\text{CosineSimilarity}(\mathbf{W}_y, \hat{f}_\theta(x_\text{eval})_\text{corrected} ))${\small\color{azure}\Comment{Predict class}}
    
\State \textbf{Output} $y_\text{pred}$

\end{algorithmic}
\end{algorithm*}

\section{Detailed analysis of ablation results (\outref{L509})}
\label{sec:appendix_ablation}
\begin{figure}[h!]

    \centering
    \begin{subfigure}[b]{1\linewidth}
        \centering
        \resizebox{0.9\linewidth}{!}{
            \includegraphics{figures/sim_legand.pdf}
        }
        \vspace{0.5em}
    \end{subfigure}
    
    \begin{subfigure}[b]{0.48\linewidth}
        \centering
        \resizebox{0.9\linewidth}{!}{
            \includegraphics{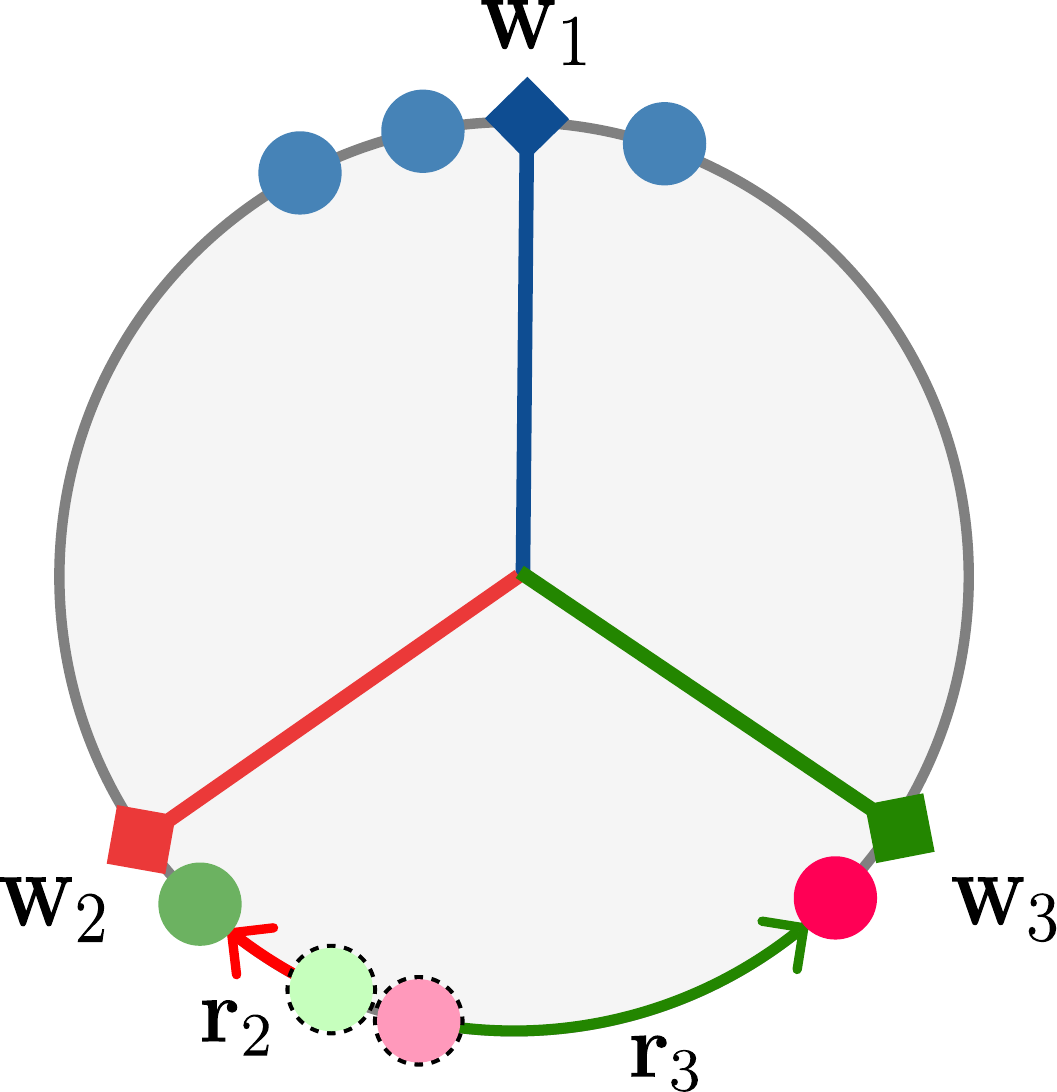}
        }
        \vspace{-0.1em}
        \caption{wrong residual added}
    \end{subfigure}
    \begin{subfigure}[b]{0.48\linewidth}
        \centering
        \resizebox{0.9\linewidth}{!}{
            \includegraphics{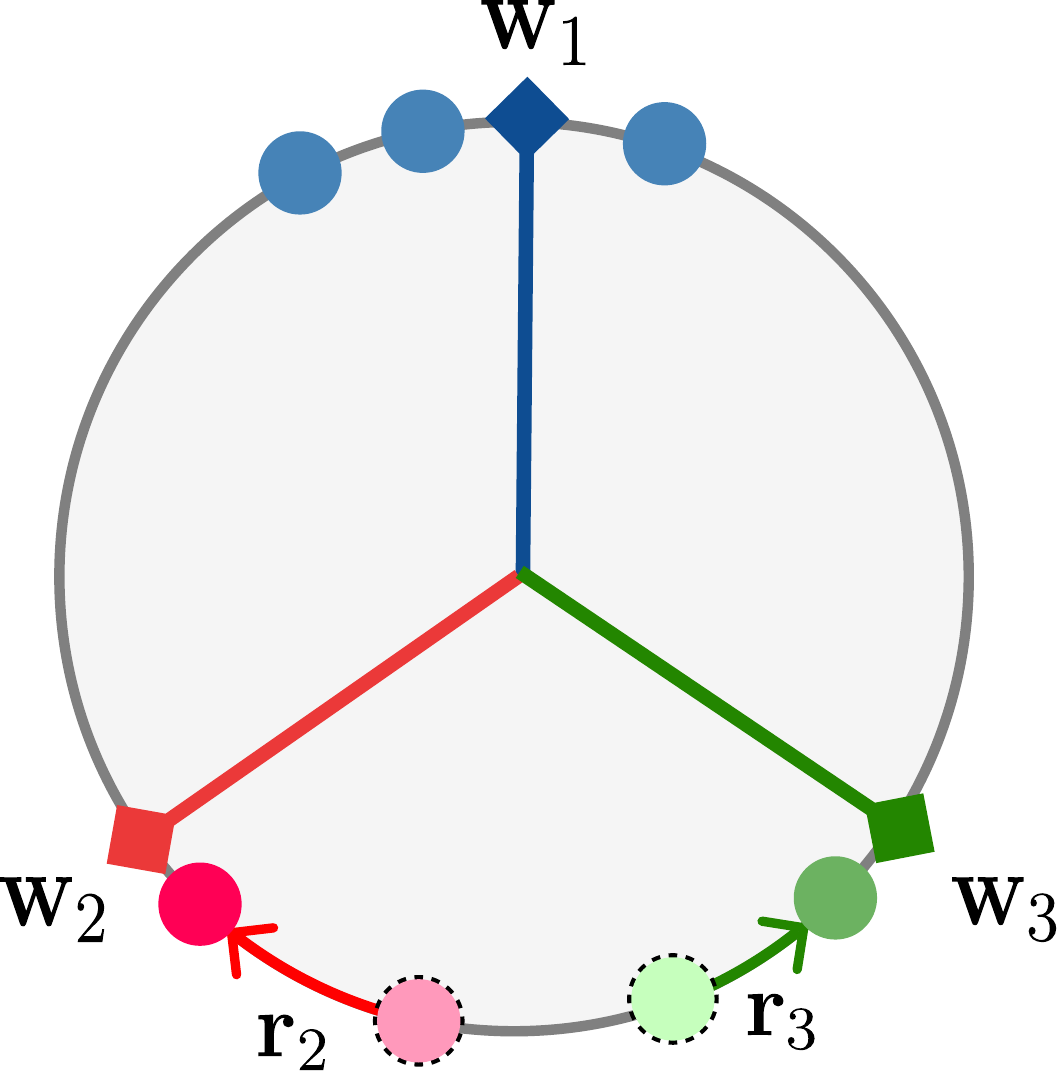}
        }
        \vspace{-0.1em}
        \caption{correct residual added}
    \end{subfigure}

    
    \caption{Without preparatory data training and relying solely on residual correction, incorrect residuals can be added due to bias, which can potentially lead to decreased performance. On the other hand, the combination of preparatory data training and joint training leads to the addition of correct residuals by addressing the bias problem.}
    \label{fig:wrong_res}
    \vspace{-.5em}
\end{figure}


Note that exclusively relying on residual correction without the training of preparatory data may result in the addition of incorrect residuals due to bias problem, as illustrated in Fig.~\ref{fig:wrong_res}-(a).
The bias in CL causes the features of novel classes and old classes to overlap.
When the features of multiple classes are clustered together, as in \outref{Fig.4-(a)}, residuals of one class can be added to the residual-correcting term of other classes in the cluster, since we select the residuals using $k$ nearest-neighbors of corresponding features.
Thus, the residuals of old classes are often added to novel class samples and vice versa, hurting the accuracy of both old and novel classes.

In this context, the use of preparatory data not only accelerates the convergence of ETF during training, but also promotes accurate residual addition during inference.
In conclusion, the combination of residual correction and preparatory data training effectively aligns the model output with the corresponding ETF classifier, as demonstrated in Fig.~\ref{fig:wrong_res}-(b).

\section{Properties of Neural Collapse}
\label{sec:nc_detail}
\textbf{(NC1) Collapse of Variability}: The last layer feature output of each data point collapses toward the class mean feature of its respective class. In other words, $h_{k,i}$, last layer feature of sample $i$ in class $k$, collapse to $\mu_k = \sum_{i=1}^{n_k}h_{k,i}$ for $\forall k\in [1,K]$ where $n_k$ is the number of samples for class $k$. 
By considering within-class covariance and between-class covariance
\begin{equation}
    \begin{split}
        \Sigma_W = \frac{1}{K}\sum_{k=1}^{K}\frac{1}{n_k}(\sum_{i=1}^{n_k}(h_{k,i} - \mu_k)), \\
        \Sigma_B = \frac{1}{K}\sum_{k=1}^{K}(\mu_k - \mu_G),    
    \end{split}
\end{equation} 
where $\mu_G = \sum_{k=1}^{K}\mu_k$, empirical variability can be measured as
\begin{equation}
    NC1 := \frac{1}{K}trace({\Sigma_W \Sigma_B^\dag}).
\end{equation}

\textbf{(NC2) Convergence to simplex equiangular tight frame (ETF)}: 
Class means $\mu_k (k \in [1,K])$ centered by the global mean $\mu_G$ converge to vertices of a simplex ETF structure, i.e., matrix $M = [m_1~m_2~\cdots~m_K]$ where $m_k=\frac{\mu_k - \mu_G}{\|\mu_k - \mu_G\|^2}$ satisfies the following equation:
\begin{equation}
    \mathbf{M}\mathbf{M}^T = \frac{1}{K-1}(K\mathit{I}_K - \mathbf{1}_K \mathbf{1}_K^T).
\end{equation}
The degree of convergence can be measured using:
\begin{equation}
    \frac{\mathbf{M}\mathbf{M}^T}{\|\mathbf{M}\mathbf{M}^T\|_F} - \frac{1}{\sqrt{K-1}}(\mathit{I}_K-\frac{1}{K}\mathbf{1}_K \mathbf{1}_K^T).
\end{equation}

\textbf{(NC3) Convergence to self-duality}:
Classifier $\mathbf{W}$ converges to the simplex ETF $\mathbf{M}$ formed by recentered feature mean, and during this convergence, the classifier vector $w_k$ aligns with their corresponding feature mean $m_k$ where $w_k$ means classifier weight for class $k$, $k \in [1,K]$, \ie,
\begin{equation}
    \frac{\textbf{M}}{~\|\textbf{M}\|_F} = \frac{\textbf{W}}{~\|\textbf{W}\|_F}
\end{equation}
Duality can be measured by measuring:
\begin{equation}
    \frac{\mathbf{W}\mathbf{M}^T}{\|\mathbf{W}\mathbf{M}^T\|_F} - \frac{1}{\sqrt{K-1}}(\mathit{I}_K-\frac{1}{K}\mathbf{1}_K \mathbf{1}_K^T).
\end{equation}




\end{document}